\documentclass[preprints,article,submit,pdftex,moreauthors,onecolumn]{Definitions/mdpi}

\usepackage[utf8]{inputenc}
\usepackage[T1]{fontenc}

\usepackage{amsmath,amssymb,amsfonts}
\usepackage{mathtools}

\usepackage{graphicx}
\usepackage{booktabs}
\usepackage{longtable}
\usepackage{array}
\usepackage{tabularx}
\usepackage{float}
\usepackage{url}
\usepackage{tikz}
\usetikzlibrary{arrows.meta, positioning, calc, decorations.pathreplacing}
\usepackage{enumitem}
\setlist[enumerate]{leftmargin=*, labelsep=0.5em}
\setlist[description]{leftmargin=1.2cm, labelwidth=1cm, labelsep=0.2cm}

\usepackage{algorithm}
\usepackage{algpseudocode}
\algrenewcommand\algorithmicrequire{\textbf{Input:}}
\algrenewcommand\algorithmicensure{\textbf{Output:}}

\firstpage{1}
\pubvolume{1}
\issuenum{1}
\articlenumber{0}
\pubyear{2026}
\copyrightyear{2025}
\datereceived{}
\dateaccepted{}
\datepublished{}

\Title{On RGB–TIR Stereo Calibration under Extreme Resolution Asymmetry}

\Author{
Michał Król $^{1}$,
Michał Salamonowicz $^{2}$,
Władysław Skarbek $^{3}$,
Michał Tomaszewski $^{4}$
}

\address{
$^{1}$ Independent Researcher, Poland; mkrol2k@gmail.com\\
$^{2}$ Warsaw University of Technology, Poland; michal.salamonowicz.dokt@pw.edu.pl\\
$^{3}$ Warsaw University of Technology, Poland; wladyslaw.skarbek@pw.edu.pl\\
$^{4}$ Polish Japanese Academy of Information Technology, Poland; tomaszew@pjwstk.edu.pl
}
\corres{Correspondence: mkrol2k@gmail.com}

\newif\ifarxiv
\arxivtrue

\begin{document}
\maketitle

\nolinenumbers

\ifarxiv
\begin{center}
\textbf{Abstract}
\end{center}
\begin{abstract}\relax
The integration of visible-light (RGB) and thermal infrared (TIR) imaging enables, for example,
three-dimensional analysis of building geometry and heat transfer phenomena, yet
the accurate geometric calibration of RGB--TIR camera systems remains challenging in
low-cost configurations with extreme resolution mismatch. This paper proposes a
stereo calibration framework for an RGB camera ($2028 \times 1520$~px) paired with
a TIR camera at only $80 \times 62$~px---a ratio of approximately $1{:}625$ in
pixel count. An active OLED-based calibration object sequentially displays modality-specific
patterns (checkerboard for TIR, ChArUco board for RGB) on the same surface, providing controlled thermal contrast that is substantially less sensitive
to ambient conditions than passive calibration objects. A dedicated thermal corner detection
algorithm based on perspective rectification, Hessian saddle-point analysis, and
Mean Shift localisation enables reliable checkerboard detection at $80 \times 62$~px
without per-frame parameter tuning, where standard detectors (e.g.\ OpenCV
\texttt{findChessboardCorners}) fail entirely. A baseline-constrained bundle adjustment
with Tz = 0 incorporates the known co-planar rig geometry, yielding physically consistent and
stable baseline estimates. The calibrated system is validated
on a thermally active building mock-up using both constant-depth and DepthPro-based
per-pixel depth estimation, demonstrating consistent TIR-to-RGB projection suitable
for building envelope analysis and geometrically consistent thermal projection onto RGB-derived building geometry.
\end{abstract}
\fi

\ifarxiv
\bigskip
\noindent\textbf{Keywords:}
RGB–thermal (TIR) camera calibration; active OLED screen; RGB–TIR data fusion; 3D scene reconstruction; depth estimation;
\fi

\section{Introduction}
\label{sec:introduction}

\subsection{Motivation and Context}

The integration of visible-spectrum (RGB) and thermal infrared (TIR) imagery is increasingly employed in engineering disciplines such as robotics, civil engineering, environmental monitoring, and geodesy. These two sensing modalities provide complementary information: RGB cameras offer high spatial resolution and rich geometric and textural detail, whereas TIR cameras capture the spatial distribution of emitted infrared radiation, enabling observation of temperature-related surface characteristics independently of illumination conditions.

The urgency of improving building energy diagnostics is underscored not only by the scale of energy consumption in the built environment, but also by the growing need for automated, large-scale assessment of building stocks. This includes the implementation of systematic, data-driven processes for detailed evaluation of groups of buildings, as well as the automated generation of preliminary retrofit recommendations. According to the International Energy Agency (IEA), the building and construction sector accounts for over 36\% of global final energy use and approximately 37\% of energy- and process-related CO$_2$ emissions~\cite{IEA_Buildings_2023}. Within the European Union, buildings are responsible for approximately 40\% of total energy consumption and 36\% of energy-related greenhouse gas emissions. The revised Energy Performance of Buildings Directive (EPBD, 2024) calls for at least doubling the annual renovation rate by 2030 and emphasises the role of advanced digital tools and data-driven analysis in achieving a decarbonised building stock by 2050. These considerations motivate the development of practical and scalable methods for multimodal building inspection, in which accurate RGB–TIR calibration constitutes an essential enabling step \cite{WolkReinhart2025}.

Accurate geometric calibration of RGB--TIR camera systems is a prerequisite for coherent multimodal analysis, as it enables the transformation of observations between the coordinate systems of the two sensors. This capability forms the foundation for data fusion, depth estimation, and three-dimensional (3D) scene reconstruction. In the context of building engineering, the combination of RGB-derived geometry with spatially aligned thermal data enables the generation of 3D thermal models that simultaneously represent the geometric form of buildings. Such models can also support the estimation of the occurrence of thermal bridges within a building, as well as the assessment of the thermal properties of its external envelope components~\cite{Dlesk2019,Zheng2020,Hoegner2018,Hassan2025,Iwaszczuk2017}. Such models provide a basis for quantifying thermal bridges, insulation defects, and patterns of energy leakage---information that is essential for reliable building energy performance assessment.

\subsection{Research Gap and Challenges}
\label{sec:research_gap}

Despite the growing interest in RGB--TIR sensing, accurate geometric calibration of such systems remains challenging, particularly when low-cost or compact thermal sensors are employed~\cite{Hoegner2018,Dlesk2019}. In this study, a configuration exhibiting an extreme resolution mismatch is considered: the RGB camera provides $2028 \times 1520$~pixels, while the TIR camera operates at only $80 \times 62$~px---a ratio of approximately $1{:}625$ in pixel count. For comparison, most existing calibration studies employ TIR sensors with resolutions of $320 \times 240$~pixels or higher~\cite{Brenner2023} or $640 \times 480$~pixels~\cite{ElSheikh2023}, and no prior work has reported \emph{RGB--TIR stereo} calibration at the resolution used here. The closest precedent, by Zoetgnand\'e et al.~\cite{Zoetgnande2019}, addresses a homogeneous \emph{thermal--thermal} stereo pair at $80 \times 60$~pixels, which avoids the cross-modality resolution imbalance (see Section~\ref{sec:related_work}).

This extreme resolution disparity introduces three interconnected challenges:

\begin{enumerate}
    \item \textbf{Unreliable feature detection in TIR imagery.} At a resolution of $80 \times 62$~px, a single checkerboard square occupies merely a few pixels at typical calibration distances, making the corner regions exceedingly small. Standard corner detection algorithms, such as \texttt{findChessboardCorners} in OpenCV, fail entirely at this resolution due to insufficient spatial support for gradient-based analysis~\cite{Hoegner2018,Dlesk2019,Placht2014}.

    \item \textbf{Sensitivity of passive and semi-active calibration objects to environmental conditions.} Unlike fully active calibration objects, passive and semi-active calibration objects (such as externally heated patterns) do not generate a tightly controlled thermal pattern; instead, contrast emerges from incidental emissivity differences, ambient heating effects, or uneven external heat application, which vary with environmental conditions and surface properties. This leads to unstable thermal contrast and poor repeatability of corner detection across acquisition sessions~\cite{ElSheikh2023,Dlesk2019}.

    \item \textbf{Manual tuning of preprocessing parameters.} Threshold values, filter settings, and normalisation schemes for thermal imagery are highly dependent on acquisition conditions and lack transferability across sessions or sensor configurations, making standardised calibration difficult~\cite{ElSheikh2023,Chen2022ThermalInfraredCalibration}.
\end{enumerate}

Recent deep learning--based approaches, such as ThermoNeRF~\cite{Hassan2025} and ThermalGS~\cite{Liu2025}, demonstrate the potential of neural scene representations for joint RGB--TIR reconstruction. However, these methods assume pre-aligned imagery obtained via Structure-from-Motion and do not address the underlying geometric calibration problem. Moreover, they require large training datasets and substantial computational resources, which limits their applicability in practical, field-deployable calibration workflows.

Consequently, there remains a need for calibration approaches that provide reliable thermal feature detection at very low TIR resolutions, operate robustly without per-frame manual tuning, and preserve the geometric interpretability of classical calibration methods.

\subsection{Contributions}
\label{sec:contributions}

This paper proposes a practical and reproducible calibration framework for RGB--TIR stereo systems exhibiting extreme resolution mismatch. The principal contributions are as follows:

\begin{enumerate}
    \item[\textbf{C1.}] An \textbf{active OLED screen} is introduced as the calibration object, which provides controlled and repeatable thermal contrast that is substantially less sensitive to ambient conditions than passive or semi-active calibration objects. Modality-specific patterns---a checkerboard for TIR and a ChArUco board for RGB---are dynamically switched on the same physical surface without moving the camera rig.

    \item[\textbf{C2.}] A \textbf{dedicated corner detection algorithm for very low-resolution thermal imagery} is proposed, combining perspective rectification, Hessian saddle-point analysis, and Mean Shift localisation. The algorithm enables reliable checkerboard detection at resolutions as low as $80 \times 62$~px without per-frame parameter tuning or learning-based detectors.

    \item[\textbf{C3.}] A \textbf{baseline-constrained bundle adjustment} is introduced to incorporate the known co-planar rig geometry, reducing ambiguity in the estimated translation and yielding physically consistent results while balancing reprojection accuracy with geometric plausibility.

    \item[\textbf{C4.}] The complete pipeline is \textbf{validated on a thermally active building mock-up}, demonstrating consistent alignment of thermal patterns in TIR-to-RGB projections. Validation encompasses both marker-based (ChArUco) physical depth and neural depth estimation (DepthPro), showcasing the utility of the approach even when explicit depth measurement is unavailable.
\end{enumerate}

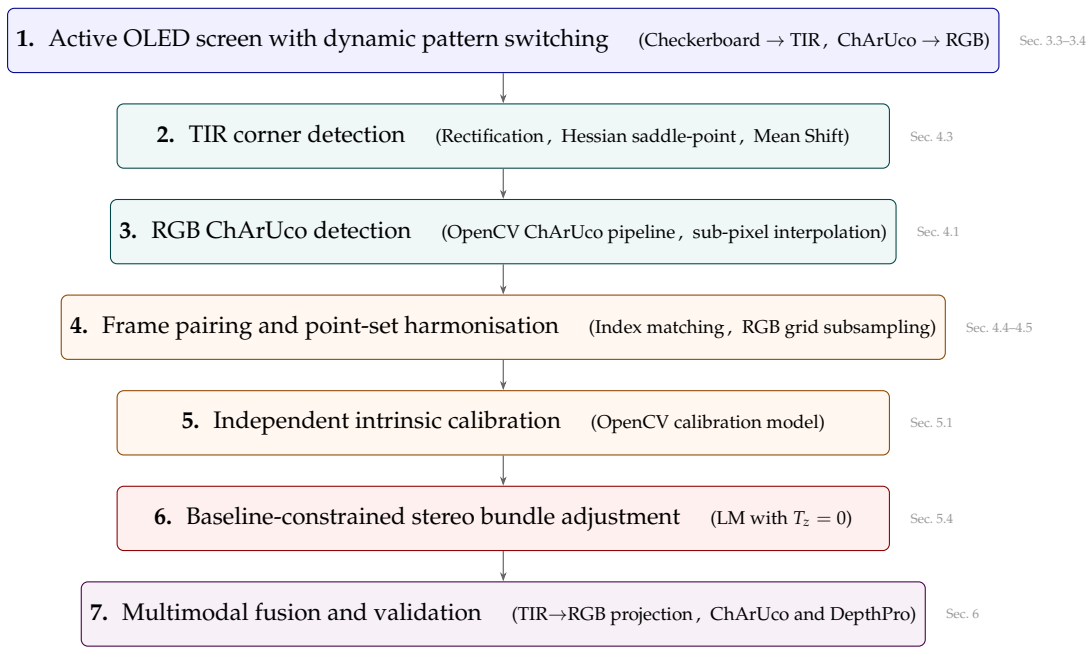
\begin{figure}[H]
\centering
\begin{tikzpicture}[
    block/.style={
        rectangle, rounded corners=2.5pt,
        minimum width=10.2cm, minimum height=0.85cm,
        align=center, font=\small,
        draw=#1!55!black, line width=0.4pt,
        fill=#1!6, text=black,
    },
    block/.default={black},
    blkB/.style={block=blue},
    blkT/.style={block=teal},
    blkO/.style={block=orange},
    blkR/.style={block=red},
    blkV/.style={block=violet},
    arr/.style={-{Stealth[length=3.5pt, width=2.5pt]}, line width=0.45pt, color=black!60},
    seclabel/.style={font=\tiny, text=black!40, anchor=west},
]

\node[blkB] (s1) {
    \textbf{1.}\; Active OLED screen with dynamic pattern switching
    \quad{\scriptsize(Checkerboard $\to$ TIR\,,\; ChArUco $\to$ RGB)}};

\node[blkT, below=0.4cm of s1] (s2) {
    \textbf{2.}\; TIR corner detection
    \quad{\scriptsize(Rectification\,,\; Hessian saddle-point\,,\; Mean Shift)}};

\node[blkT, below=0.4cm of s2] (s3) {
    \textbf{3.}\; RGB ChArUco detection
    \quad{\scriptsize(OpenCV ChArUco pipeline\,,\; sub-pixel interpolation)}};

\node[blkO, below=0.4cm of s3] (s4) {
    \textbf{4.}\; Frame pairing and point-set harmonisation
    \quad{\scriptsize(Index matching\,,\; RGB grid subsampling)}};

\node[blkO, below=0.4cm of s4] (s5) {
    \textbf{5.}\; Independent intrinsic calibration
    \quad{\scriptsize(OpenCV calibration model)}};

\node[blkR, below=0.4cm of s5] (s6) {
    \textbf{6.}\; Baseline-constrained stereo bundle adjustment
    \quad{\scriptsize(LM with $T_z = 0$)}};

\node[blkV, below=0.4cm of s6] (s7) {
    \textbf{7.}\; Multimodal fusion and validation
    \quad{\scriptsize(TIR$\to$RGB projection\,,\; ChArUco and DepthPro)}};

\foreach \i/\j in {s1/s2, s2/s3, s3/s4, s4/s5, s5/s6, s6/s7}{
    \draw[arr] (\i.south) -- (\j.north);
}

\node[seclabel] at ($(s1.east) + (0.15,0)$) {Sec.~3.3--3.4};
\node[seclabel] at ($(s2.east) + (0.15,0)$) {Sec.~4.3};
\node[seclabel] at ($(s3.east) + (0.15,0)$) {Sec.~4.1};
\node[seclabel] at ($(s4.east) + (0.15,0)$) {Sec.~4.4--4.5};
\node[seclabel] at ($(s5.east) + (0.15,0)$) {Sec.~5.1};
\node[seclabel] at ($(s6.east) + (0.15,0)$) {Sec.~5.4};
\node[seclabel] at ($(s7.east) + (0.15,0)$) {Sec.~6};

\end{tikzpicture}
\caption{Overview of the proposed RGB--TIR stereo calibration pipeline. Each stage corresponds to a dedicated section of the paper, as indicated on the right.}
\label{fig:pipeline_overview}
\end{figure}

All experiments were conducted using a low-cost, self-assembled hardware setup. The overall calibration workflow is illustrated schematically in Figure~\ref{fig:pipeline_overview}.

\section{Related Work}
\label{sec:related_work}

Research directly relevant to this work spans three themes: (1)~geometric calibration of RGB--thermal infrared (TIR) camera systems, with particular attention to configurations exhibiting resolution mismatch; (2)~multimodal 3D reconstruction of building envelopes; and (3)~corner detection in very low-resolution thermal imagery. Each subsection is organised around the specific gap that the present work addresses.

\subsection{Geometric Calibration of RGB--TIR Systems}
\label{sec:rw_calibration}

Camera calibration is grounded in the pinhole model and the planar-calibration-object formulation of Zhang~\cite{Zhang2000}, which is the de facto standard for RGB cameras through implementations such as OpenCV~\cite{OpenCV_Calib3d}. Transferring this framework to the thermal domain is not straightforward: TIR image formation depends on emissivity, temperature, and ambient conditions rather than on reflected visible light, and low-cost TIR sensors exhibit reduced spatial resolution, lower contrast, and elevated noise compared with their visible-spectrum counterparts~\cite{ElSheikh2023}. Existing approaches to RGB--TIR calibration can be grouped into passive-calibration-object, active-calibration-object, and target-free methods; below, we discuss representative works along the axis that matters most for the present study---the ability to produce reliable geometric constraints at very low TIR resolution.

\paragraph{\textit{Passive and heated calibration objects}}
The majority of RGB--TIR calibration methods rely on passive or statically heated calibration objects, whose thermal visibility arises from material emissivity differences or uniform external heating. Alba et al.~\cite{Alba2011} introduced a procedure for texturing thermal IR images onto a 3D building model acquired by terrestrial laser scanning, using calibration objects such as wooden panels with iron nails; ElSheikh et al.~\cite{ElSheikh2023} presented a comprehensive review of passive thermal checkerboards and proposed a precise thermal radiation calibration object, demonstrating that emissivity contrast and surface material properties critically influence intrinsic parameter estimation. Saponaro et al.~\cite{Saponaro2015} improved stereo calibration of thermal cameras ($640 \times 480$~px) by heating a paper-printed checkerboard target placed on a ceramic tile, exploiting the tile's thermal mass to sustain contrast over time. Roshan et al.~\cite{Roshan2024} used a ChArUco board machined from an aluminium sheet and heated externally to calibrate a $320 \times 256$~px thermal camera, with corners extracted via the OpenCV \texttt{detectMarkers} and \texttt{interpolateCornersCharuco} functions. Vidas et al.~\cite{Vidas2012} moved a step towards controllable contrast by placing a cardboard mask in front of an emissive backdrop (a powered computer monitor or warm laptop case) whose thermal radiance provides the necessary gradient without external heating; however, the backdrop pattern is fixed and its emission is not spatially programmable. In the building inspection context closest to ours, Sher et al.~\cite{Sher2023} evaluated three board constructions (cardboard--acrylic, wood--vinyl, and metal--vinyl) with both checkerboard and ChArUco geometries for calibrating a thermal--RGB pair used in UAV-based envelope inspection. The recurring limitation of this family of methods is that thermal contrast is ambient-dependent and decays over time as calibration objects thermally equilibrate with their surroundings~\cite{ElSheikh2023,Piccinelli2024}, motivating the search for active calibration objects that can produce controlled and repeatable contrast under laboratory conditions. The present work addresses this limitation through an OLED screen that provides per-pixel emission control and dynamic pattern switching (contribution C1).

\paragraph{\textit{Active and emissive calibration objects}}
To stabilise thermal contrast, Chen et al.~\cite{Chen2022ThermalInfraredCalibration} investigated actively heated calibration patterns and showed that they significantly improve calibration stability in thermal--visual systems relative to passive boards. However, their active heating is applied uniformly to a fixed geometric pattern and does not permit modification of the displayed pattern or independent per-pixel control of emission. Our work extends the active-calibration-object paradigm by employing a programmable OLED screen, which allows both modalities to be presented with their own optimal pattern on the same physical surface without moving the rig (Section~\ref{sec:pattern_switching}).

\paragraph{\textit{Target-free and pose-based registration}}
An alternative line of work avoids direct feature detection in TIR imagery by leveraging geometry obtained from other modalities. Vidas et al.~\cite{Vidas2013} proposed a line-based target-free calibration between thermal and depth-derived features. Hoegner et al.~\cite{Hoegner2018} exploited 2D--3D correspondences between TIR images and an RGB-derived 3D point cloud to refine the exterior orientation of a multi-sensor platform. Elias et al.~\cite{Elias2023} replaced hand-crafted features with deep feature matching (SuperGlue) followed by PnP pose estimation for aligning thermal imagery with RGB-derived point clouds. These approaches reduce reliance on direct thermal feature detection, but they transfer the burden onto the RGB-derived geometry and onto scene-specific structures (lines, textured surfaces), which are typically unavailable in the controlled, planar-calibration-object scenario addressed here.

\paragraph{\textit{Stereo calibration at low TIR resolution}}
Two recent studies most directly anticipate the problem we solve. Piccinelli et al.~\cite{Piccinelli2024} proposed a passive stereo calibration technique for visible--thermal imaging systems operating at low TIR resolution, with a dedicated feature-extraction pipeline robust to specular reflections and contrast inversion. Zoetgnand\'e et al.~\cite{Zoetgnande2019} calibrated a stereo pair of $80 \times 60$~pixel thermal cameras using sub-pixel analysis and bootstrap methods, achieving quarter-pixel reprojection error and outperforming standard OpenCV stereo calibration at that resolution. Zoetgnand\'e et al.\ work at the same resolution regime considered here, but their system is \emph{thermal--thermal}, so both views share the same image characteristics and the planar-calibration-object degeneracy that we address does not arise. Piccinelli et al.\ couple low-resolution TIR with a high-resolution RGB camera---closest to our configuration---yet rely on a passive calibration object and do not report an explicit resolution ratio comparable to the $1{:}625$ considered here. Neither work combines an active OLED screen, a dedicated detector for extreme-low-resolution TIR, and a constrained bundle adjustment that enforces physically meaningful rig geometry; the present paper does.

\paragraph{\textit{Relation to our prior work}}
In a closely related study using the same OLED screen and acquisition rig, Skarbek et al.~\cite{Skarbek2026} investigated monocular pose estimation of RGB and TIR cameras using the PnP-ProCay78 algorithm with Cayley rotation parameterisation. That work estimates extrinsic parameters of each camera independently; it does not address intrinsic calibration, stereo geometry, or robust thermal corner detection. The present paper builds on that foundation and contributes the missing stereo component: dedicated TIR corner detection (Section~\ref{sec:ir_detector}), cross-modality point-set harmonisation (Section~\ref{sec:harmonisation}), and baseline-constrained bundle adjustment (Section~\ref{sec:constrained_ba}).

\subsection{Multimodal 3D Reconstruction for Building Envelope Analysis}
\label{sec:rw_reconstruction}

The immediate application of accurate RGB--TIR calibration is the generation of 3D thermal models representing both building geometry and surface heat transfer. Classical photogrammetric workflows combine RGB or laser-scanning-derived geometry with projected thermal imagery: Iwaszczuk and Stilla~\cite{Iwaszczuk2017} refined camera pose by matching uncertain 3D building models to TIR image sequences to enable high-quality thermal texture extraction, while Hoegner et al.~\cite{Hoegner2018} fused TIR images with dense 3D point clouds of indoor scenes. Dlesk and Vach~\cite{Dlesk2019} reconstructed 3D point clouds directly from TIR imagery using Structure-from-Motion with passive control points, achieving a check-point RMSE of approximately $19$~mm while noting that low resolution and weak texture of TIR images limit reconstruction quality---a constraint that resonates directly with the configuration considered in this paper.

More recent work has moved towards neural scene representations. Hassan et al.~\cite{Hassan2025} proposed ThermoNeRF, a multimodal neural radiance field for joint RGB--thermal view synthesis of building facades with a reported mean absolute temperature error of approximately $1.5\,^{\circ}\mathrm{C}$. Parallel efforts extend thermal scene modelling to 3D Gaussian Splatting~\cite{Liu2025,Lu2025}. All of these methods rely on camera poses estimated from RGB imagery via Structure-from-Motion (e.g.\ COLMAP) and assume that the RGB and TIR modalities are either hardware-aligned or provided as pre-registered pairs; they do not address the underlying geometric calibration problem. The calibration pipeline proposed here is precisely the step that such reconstruction methods presuppose but do not themselves provide.

\subsection{Corner Detection in Low-Resolution Thermal Imagery}
\label{sec:rw_detection}

Reliable corner detection in thermal images is the practical bottleneck of any calibration pipeline that uses calibration-object-based patterns. Unlike RGB imagery, thermal image intensity is a nonlinear function of emissivity and temperature rather than reflected illumination, producing non-stationary intensity profiles that violate the gradient-constancy assumptions underlying standard detectors~\cite{ElSheikh2023,Chen2022ThermalInfraredCalibration}. At moderate resolutions ($320 \times 240$~pixels and above), standard checkerboard detectors can be applied with appropriate preprocessing~\cite{ElSheikh2023}. As resolution decreases, however, the spatial support around each corner becomes insufficient for reliable gradient estimation, and dedicated strategies are required.

Placht et al.~\cite{Placht2014} demonstrated that OpenCV's \texttt{findChessboardCorners} fails under extreme poses and low-resolution conditions, and proposed the ROCHADE detector, which exploits the saddle-point geometry of checkerboard corners through the Hessian determinant. This saddle-point property, originally formalised by Beaudet~\cite{Beaudet1978}, is central to the detector proposed in Section~\ref{sec:ir_detector}. More specifically to the low-resolution regime considered here, Zoetgnand\'e et al.~\cite{Zoetgnande2019} combined sub-pixel phase analysis with bootstrap methods to calibrate an $80 \times 60$~pixel thermal stereo pair, confirming that at this resolution standard pipelines must be replaced by resolution-aware strategies. Piccinelli et al.~\cite{Piccinelli2024} took a complementary approach, replacing corner detection altogether with MSER blob centroid estimation and an iterative pattern-filling procedure, achieving robust feature extraction under specular reflections and contrast inversion in low-resolution LWIR imagery. Elias et al.~\cite{Elias2023} further observed that co-registration errors in thermal datasets frequently originate in the feature detection stage, underscoring the importance of robust and repeatable corner localisation.

No prior work, to our knowledge, reports reliable checkerboard corner detection at $80 \times 62$~px within an \emph{RGB--TIR} stereo pipeline using an \emph{active} OLED screen. The present paper closes this gap through a dedicated deterministic detector (Section~\ref{sec:ir_detector}) that combines perspective rectification, Hessian saddle-point response, and Mean Shift localisation, and is gated by a frame-level geometric consistency test.

\section{Experimental Setup and Data Acquisition}
\label{sec:setup}

\subsection{Hardware Configuration}
\label{sec:hardware}

The experiments were conducted using a rigid stereo rig consisting of an RGB camera (Arducam IMX477, $2028 \times 1520$~pixels) and a TIR camera (Waveshare Thermal Camera HAT, $80 \times 62$~px, $45^{\circ}$~FOV) mounted in a fixed stereoscopic configuration with an inter-camera baseline of approximately 30~mm. Both cameras were controlled by a Raspberry Pi~5 computing unit. The rig was attached to a photographic tripod to eliminate vibrations during image capture. The two cameras recorded images sequentially without hardware-based temporal synchronisation; geometric consistency was ensured by keeping the rig stationary during each image pair acquisition.

An LG OLED 42C27LA display (42'', 4K, 120~Hz) served as the active OLED screen. OLED technology employs self-emissive organic diodes with per-pixel luminance control, producing measurable temperature contrasts between black and white pattern regions in the TIR image. The technical parameters of all hardware components are summarised in Table~\ref{tab:hardware-rgb-ir}. The setup combines a high-resolution RGB sensor with an extremely low-resolution thermal sensor, resulting in an image resolution ratio of approximately $1{:}625$ in pixel count, which constitutes one of the principal challenges addressed in this work.

For the purpose of validating the calibrated system in a realistic building inspection scenario, a scaled building mock-up (1:10) was constructed using materials commonly found in real buildings, including wood, thin-layer fa\c{c}ade plaster, mineral wool insulation, natural red brick, bituminous roofing membrane, and glass. The mock-up was equipped with an internal heating element capable of raising the interior temperature to approximately $50\,^{\circ}\mathrm{C}$.

\begin{figure}[H]
\centering
\includegraphics[width=1.0\linewidth]{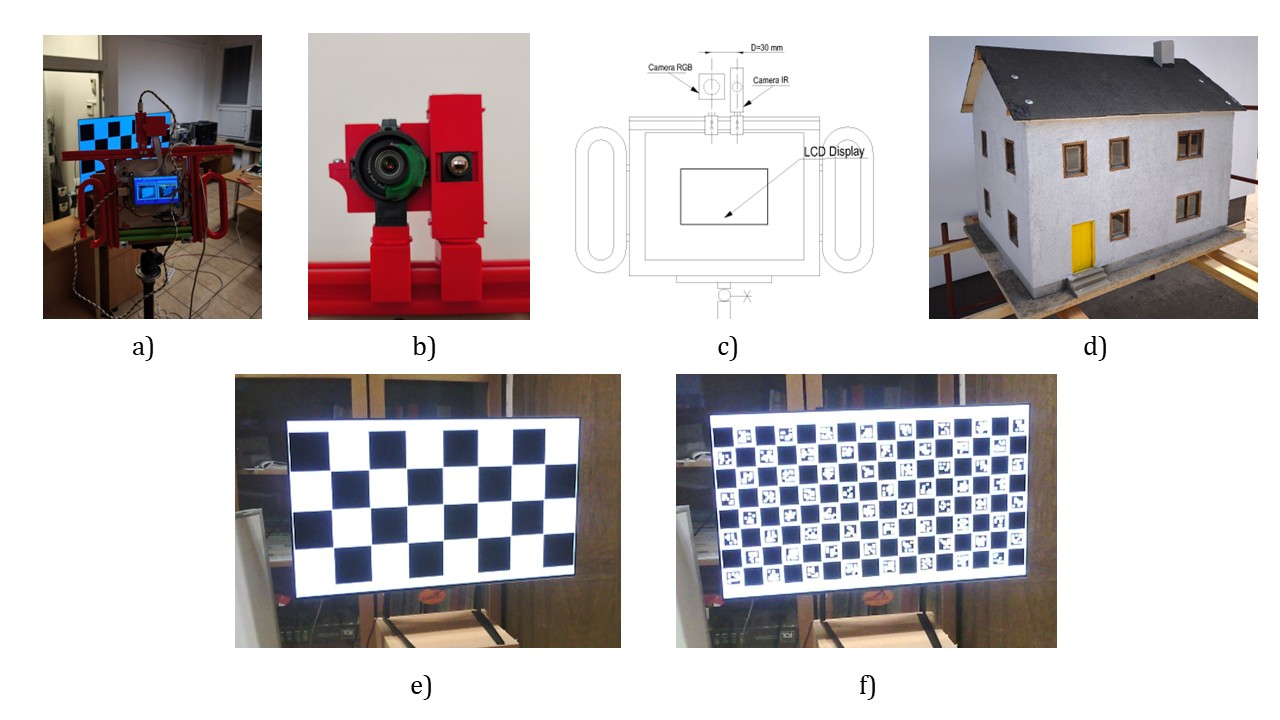}
\caption{RGB--TIR measurement setup and active OLED screen:
(a)~complete RGB--TIR acquisition rig with preview display;
(b)~front view of the rigid camera stereo configuration;
(c)~schematic of the measurement setup;
(d)~thermally active building mock-up, 1:10 scale;
(e)~checkerboard pattern displayed on the OLED screen for the TIR modality;
(f)~ChArUco pattern displayed on the OLED screen for the RGB modality.}
\label{fig:acquisition_setup}
\end{figure}

The RGB--TIR acquisition setup, building mock-up, and active OLED screen are shown in Figure~\ref{fig:acquisition_setup}.

{\scriptsize
\setlength{\LTpre}{6pt}
\setlength{\LTpost}{6pt}
\setlength{\tabcolsep}{1.8pt}
\renewcommand{\arraystretch}{1.45}

\begin{longtable}{|
>{\raggedright\arraybackslash}p{2.2cm}|
>{\raggedright\arraybackslash}p{3.2cm}|
>{\raggedright\arraybackslash}p{7.5cm}|}
\caption{Hardware components of the RGB--TIR acquisition system.}
\label{tab:hardware-rgb-ir} \\

\hline
\textbf{Component} &
\textbf{Model / Specification} &
\textbf{Description} \\
\hline
\endfirsthead

\hline
\textbf{Component} &
\textbf{Model / Specification} &
\textbf{Description} \\
\hline
\endhead

\hline
\endfoot
\hline
\endlastfoot

TIR camera &
Waveshare HAT (No.~25288), 80$\times$62~px, 45$^\circ$~FOV, USB-C &
Low-resolution infrared sensor for thermal imaging \\
\hline

RGB camera &
Arducam IMX477 HQ, 12.3~MP (B024001) &
High-resolution visible-light camera with C--CS adapter and tripod mount \\
\hline

RGB lens &
M12 8~mm fixed-focus (LN024) &
Fixed-focus lens providing low optical radial distortion for calibration scenes \\
\hline

Computing unit &
Raspberry Pi~5, 8~GB RAM &
Controls both cameras and transmits acquired data \\
\hline

Calibration display &
LG OLED 42C27LA, 42'', 4K, 120~Hz &
Self-emissive OLED display providing per-pixel thermal contrast for TIR calibration \\
\hline

Mounting rig &
Custom 3D-printed holder (FDM) &
Rigid fixture for both cameras with ${\sim}30$~mm baseline, attachable to a tripod \\
\hline

\end{longtable}
}

\subsection{Active OLED Calibration Screen}
\label{sec:oled_target}

A key requirement for calibrating very low-resolution TIR cameras is a calibration object that produces sufficient and repeatable thermal contrast. Passive calibration boards, whose thermal visibility relies on differences in material emissivity or uncontrolled heating, are known to yield unstable contrast that varies with ambient conditions~\cite{ElSheikh2023,Chen2022ThermalInfraredCalibration}.

\begin{figure}[H]
\centering
\includegraphics[width=1.0\linewidth]{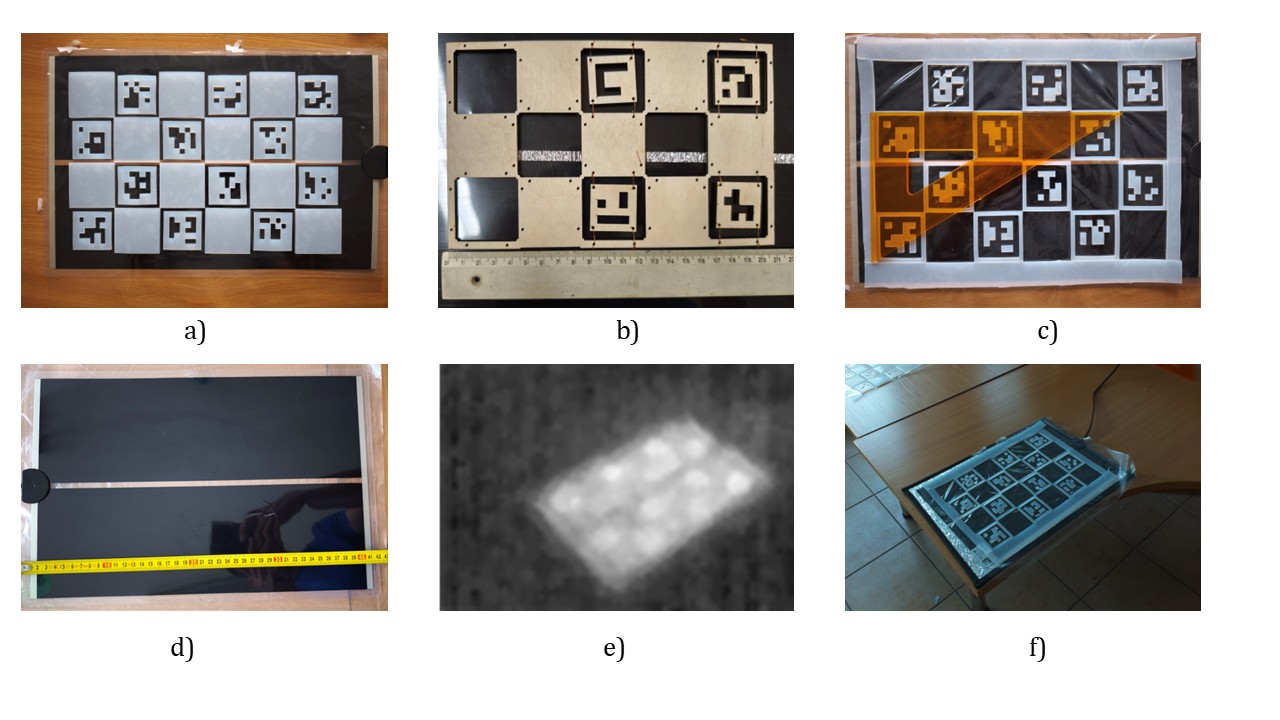}
\caption{Preliminary passive and semi-active calibration objects evaluated during initial experiments:
(a--f)~examples of silicone and plywood ChArUco boards and further passive calibration object variants.}
\label{fig:preliminary_targets}
\end{figure}

To investigate alternative calibration object designs, preliminary experiments were conducted with two types of semi-active calibration objects prior to adopting the OLED-based solution (Figure~\ref{fig:preliminary_targets}). First, a calibration board was fabricated from 3~mm thick silicone with ChArUco patterns cut using a CNC machine. Silicone was selected for its low thermal conductivity, which promotes contrast formation in TIR imagery. The board was heated using a terrarium-grade electric heating mat maintaining a surface temperature of approximately $27\,^{\circ}\mathrm{C}$. Second, a similar experiment was performed using a ChArUco pattern laser-cut from 3~mm thick plywood. Both calibration objects exhibited non-uniform heating, imperfect pattern edges, and insufficient thermal contrast for reliable corner detection at $80 \times 62$~px. Moreover, these physical calibration objects did not permit automated pattern switching or precise spatial alignment of calibration patterns across modalities.

Based on these findings, an OLED display was adopted as the active OLED screen. Light emission from the display induces localised temperature variations across the pattern elements, producing thermal contrast between adjacent black and white regions that is directly observable in the TIR image. This mechanism yields repeatable contrast that is substantially less sensitive to ambient temperature fluctuations than that obtained using passive or semi-active calibration objects.

An additional advantage of the OLED-based solution lies in its flexibility: the displayed calibration pattern can be modified programmatically in terms of geometry, scale, and visual parameters without any changes to the physical setup. This enables rapid testing of different pattern configurations and straightforward adaptation to the characteristics of each sensing modality.

\subsection{Dynamic Pattern Switching and Data Acquisition Protocol}
\label{sec:pattern_switching}

A single calibration pattern is rarely optimal for both the RGB and TIR modalities simultaneously. The TIR camera requires a pattern that provides strong thermal contrast with a geometry resolvable at very low resolution, whereas the RGB camera benefits from patterns that enable unambiguous corner identification and precise sub-pixel localisation. To address this, modality-specific patterns are employed: a classical checkerboard with parameters $(S_x, S_y) = (8, 4)$ for the TIR camera, and a ChArUco board with parameters $(S_x, S_y) = (16, 8)$ and marker dictionary \texttt{DICT\_5X5\_100} for the RGB camera~\cite{GarridoJurado2014,OpenCV_Charuco}.

During data acquisition, the calibration patterns are switched dynamically on the OLED display without altering the position of the camera rig. The acquisition sequence proceeds as follows: (1)~the checkerboard pattern is displayed and the OLED surface is allowed to reach thermal equilibrium (approximately 10~s heating phase; this duration was determined empirically by direct visual inspection of the live thermal preview described in Section~\ref{sec:hardware}, with image capture triggered once no further improvement in checkerboard contrast was observable); (2)~the TIR image is captured; (3)~the display is switched to the ChArUco pattern (switching delay: 0.1~s); (4)~the corresponding RGB image is captured. The short switching delay is sufficient for the RGB modality, which relies on luminance rather than thermal contrast. Pattern display is controlled remotely via a lightweight HTTP server, providing deterministic and repeatable operation without manual intervention.

It should be noted that the thermal contrast of the OLED display varied over time during prolonged operation: contrast was initially low, increased as the display warmed up, and gradually decreased after extended heating. This behaviour highlights the importance of controlled acquisition timing when using actively emitting calibration objects.

Calibration data were collected during four acquisition sessions (Sessions~I--IV) conducted under laboratory conditions (ambient temperature $21 \pm 2\,^{\circ}\mathrm{C}$). Sessions~I and~II were exploratory in nature and served to develop and refine the data acquisition protocol, including the OLED warm-up timing, camera positioning strategy, and the thermal corner detection pipeline. Although these sessions yielded usable RGB data, the stereo results were unsatisfactory due to insufficient thermal contrast control and a limited number of valid TIR detections. The quantitative results reported in this paper are based on Sessions~III and~IV, which followed the finalised protocol and produced the most complete and consistent datasets. In each of these sessions, the camera rig was positioned at distances ranging from approximately 1 to 3~m from the OLED display, covering a range of orientations including frontal views and large inclination angles along two axes. Between 40 and 60 image pairs were recorded per session.

\section{Detection and Matching of Calibration Points}
\label{sec:detection_matching}

\subsection{RGB Point Detection}

Calibration point detection in the RGB modality is performed using the standard ChArUco marker detection and corner interpolation pipeline available in OpenCV~\cite{GarridoJurado2014,OpenCV_Charuco}. The combination of high image resolution ($2028 \times 1520$~pixels) and the use of a ChArUco calibration pattern results in stable and reliable point detection with sub-pixel localisation accuracy. Typically, between 90 and 105 corners were detected per frame, with a detection success rate exceeding 98\% across all acquisition sessions. The obtained detections exhibit high repeatability across consecutive frames and do not require modality-specific preprocessing.

\subsection{Limitations of Standard Corner Detection in Thermal Imagery}
\label{sec:ir_limitations}

As discussed in Section~\ref{sec:research_gap}, the detection of calibration points in thermal infrared images at $80 \times 62$~px resolution constitutes a substantially more challenging problem than in the RGB modality. The following experiments confirmed and quantified these limitations for the specific sensor and calibration object configuration used in this study.

Preliminary experiments confirmed that the standard OpenCV \texttt{findChessboardCorners} function failed to detect any corners in thermal images of this resolution, regardless of the preprocessing strategy applied. This result is consistent with the known limitations of the algorithm, which relies on connected quadrilateral detection and adaptive thresholding---operations that require sufficient spatial support around each corner to resolve the local intensity pattern. At $80 \times 62$~px, this support is fundamentally insufficient~\cite{Hoegner2018,Dlesk2019}, in line with the observations of Placht et al.~\cite{Placht2014} and the resolution-aware strategies surveyed in Section~\ref{sec:rw_detection}.

Furthermore, the effectiveness of alternative preprocessing strategies --- including bilateral filtering, non-local means denoising, contrast-limited adaptive histogram equalisation (CLAHE) \cite{Zuiderveld1994}, and various thresholding schemes --- was found to be highly sensitive to acquisition conditions. Parameter settings that enabled partial detection in one frame frequently produced no detections or false positives in another, even within the same acquisition session. This instability arises from the non-stationary thermal contrast of the OLED screen, ambient temperature variations, and the elevated noise floor of the thermal sensor \cite{ElSheikh2023}.

These findings motivated the development of a dedicated corner detection algorithm for thermal imagery that does not rely on \texttt{findChessboardCorners} or on per-frame parameter tuning, but instead exploits the known geometric structure of the calibration pattern through perspective rectification and differential image analysis.

\subsection{Proposed Thermal Corner Detection Algorithm}
\label{sec:ir_detector}

\begin{figure}[H]
\centering
\resizebox{\linewidth}{!}{
\begin{tikzpicture}[
    stage/.style={
        rectangle, rounded corners=2.5pt,
        minimum width=1.55cm, minimum height=1.05cm,
        align=center, font=\scriptsize,
        draw=#1!60!black, line width=0.4pt,
        fill=#1!8, text=black,
    },
    stage/.default={black},
    stageBlue/.style={stage=blue},
    stageRed/.style={stage=red},
    stageAmber/.style={stage=orange},
    arr/.style={-{Stealth[length=3pt, width=2.5pt]}, line width=0.45pt, color=black!70},
    grp/.style={decorate, decoration={brace, amplitude=3pt, mirror}, line width=0.35pt, black!50},
    grplabel/.style={font=\tiny\itshape, text=black!60},
    rejlabel/.style={font=\tiny\bfseries, text=red!70!black},
]

\node[stageBlue]  (s1) {\textbf{Stage 1}\\[0.5pt] Percentile\\[-1pt] normalisation};
\node[stageBlue,  right=0.3cm of s1] (s2) {\textbf{Stage 2}\\[0.5pt] ROI\\[-1pt] estimation};
\node[stageBlue,  right=0.3cm of s2] (s3) {\textbf{Stage 3}\\[0.5pt] Perspective\\[-1pt] rectification};
\node[stageBlue,  right=0.3cm of s3] (s4) {\textbf{Stage 4}\\[0.5pt] CLAHE};
\node[stageRed,   right=0.3cm of s4] (s5) {\textbf{Stage 5}\\[0.5pt] Hessian saddle\\[-1pt] response};
\node[stageRed,   right=0.3cm of s5] (s6) {\textbf{Stage 6}\\[0.5pt] Mean Shift\\[-1pt] localisation};
\node[stageAmber, right=0.3cm of s6] (s7) {\textbf{Stage 7}\\[0.5pt] Quality\\[-1pt] gate};
\node[stageBlue,  right=0.3cm of s7] (s8) {\textbf{Stage 8}\\[0.5pt] Perspective\\[-1pt] projection};

\foreach \i/\j in {s1/s2, s2/s3, s3/s4, s4/s5, s5/s6, s6/s7}{
    \draw[arr] (\i.east) -- (\j.west);
}
\draw[arr] (s7.east) -- (s8.west) node[midway, above, font=\tiny\itshape, text=black!50] {accept};
\draw[arr, red!70!black] (s7.south) -- ++(0, -0.5) node[below, rejlabel] {reject};

\draw[grp] ($(s1.south west) + (0, -0.08)$) -- ($(s4.south east) + (0, -0.08)$)
    node[midway, below=4pt, grplabel] {Preprocessing};
\draw[grp] ($(s5.south west) + (0, -0.08)$) -- ($(s6.south east) + (0, -0.08)$)
    node[midway, below=4pt, grplabel] {Corner detection};
\draw[grp] ($(s7.south west) + (0, -0.08)$) -- ($(s8.south east) + (0, -0.08)$)
    node[midway, below=4pt, grplabel] {Validation \& output};

\end{tikzpicture}
}
\caption{Overview of the proposed thermal corner detection pipeline. Stages~1--4 perform image preprocessing and geometric normalisation; Stages~5--6 detect candidate corners using differential analysis; Stage~7 evaluates geometric consistency and may reject the entire frame; Stage~8 maps accepted corners to the original image coordinates via perspective projection.}
\label{fig:ir_pipeline_overview}
\end{figure}
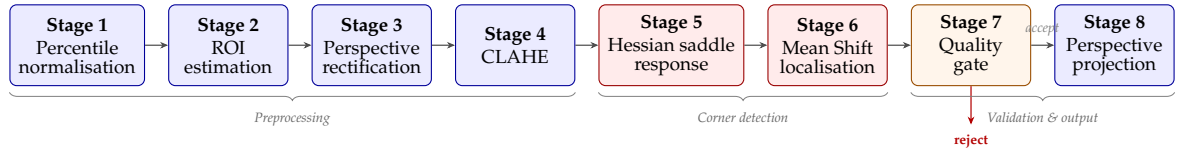

The proposed detector operates directly on raw thermal images and produces a set of checkerboard corner locations with associated grid indices, suitable for subsequent intrinsic and stereo calibration. The algorithm consists of eight sequential stages: percentile normalisation, region-of-interest (ROI) estimation, perspective rectification, contrast-limited adaptive histogram equalisation (CLAHE), Hessian saddle-point response computation, Mean Shift corner localisation, a multi-criterion quality gate, and perspective projection to the original image coordinates. The complete pipeline is deterministic and requires no per-frame parameter optimisation. An overview of the processing stages is presented in Figure~\ref{fig:ir_pipeline_overview}.

\subsubsection{Percentile Normalisation}

Raw thermal images are treated as real-valued scalar fields $I : \Omega \subset \mathbb{Z}^2 \to \mathbb{R}$. Standard min--max normalisation is sensitive to isolated extreme pixel values caused by sensor noise or hot pixels, which distort the effective dynamic range. To mitigate this effect, robust percentile normalisation is applied:
\begin{equation}
\tilde{I}(x,y) = \operatorname{clip}\!\left(\frac{I(x,y) - q_1}{q_{99} - q_1},\; 0,\; 1\right),
\label{eq:percentile_norm}
\end{equation}
where $q_1$ and $q_{99}$ denote the $1^{\mathrm{st}}$ and $99^{\mathrm{th}}$ percentiles of the image intensity distribution, respectively. This transformation maps approximately 98\% of the pixel intensities to the full $[0,1]$ range while suppressing the influence of outlier values. The normalised image $\tilde{I}$ defined by Eq.~\eqref{eq:percentile_norm} serves as the input to the subsequent ROI estimation stage.

\subsubsection{ROI Estimation and Perspective Rectification}
\label{sec:roi_rectification}

The checkerboard region within the normalised image is identified using Otsu thresholding \cite{Otsu1979}, which determines an optimal binarisation threshold $T^*$ by maximising the inter-class variance of the pixel intensity distribution. The resulting binary mask is refined through morphological closing ($5 \times 5$ kernel, two iterations) and opening ($5 \times 5$ kernel, one iteration) to remove noise-induced artefacts. The largest connected component is extracted, its convex hull is computed, and the hull contour is approximated by a quadrilateral $Q = [\mathbf{q}_0, \mathbf{q}_1, \mathbf{q}_2, \mathbf{q}_3]$ (top-left, top-right, bottom-right, bottom-left) using iterative polygon simplification with progressively increasing tolerance.

The per-axis pitch of the checkerboard is estimated directly from the quadrilateral geometry:
\begin{equation}
p_x = \frac{\bar{w}}{S_x}, \qquad p_y = \frac{\bar{h}}{S_y},
\label{eq:pitch}
\end{equation}
where $\bar{w}$ and $\bar{h}$ are the average horizontal and vertical side lengths of $Q$, and $S_x$, $S_y$ denote the number of squares along each axis. The pitch estimates from Eq.~\eqref{eq:pitch} determine both the rectified grid geometry and the scale of the subsequent differential operators.

A homography $H \in \mathbb{R}^{3 \times 3}$ mapping $Q$ to a destination rectangle of size $\bar{w} \times \bar{h}$ is computed, and the normalised image is warped accordingly. Contrast-limited adaptive histogram equalisation (CLAHE) \cite{Zuiderveld1994} is then applied to the warped image at native resolution, followed by bicubic upsampling by a factor of $\kappa = 4$. Performing CLAHE before upsampling ensures that local contrast enhancement operates on the original sensor data, and the subsequent interpolation smooths any tile-boundary artefacts. The effective pitch in the upscaled rectified image is $p_x^* = \kappa \, p_x$ and $p_y^* = \kappa \, p_y$, providing increased spatial support for the subsequent differential analysis.

\begin{figure}[H]
\centering
\includegraphics[width=0.95\linewidth]{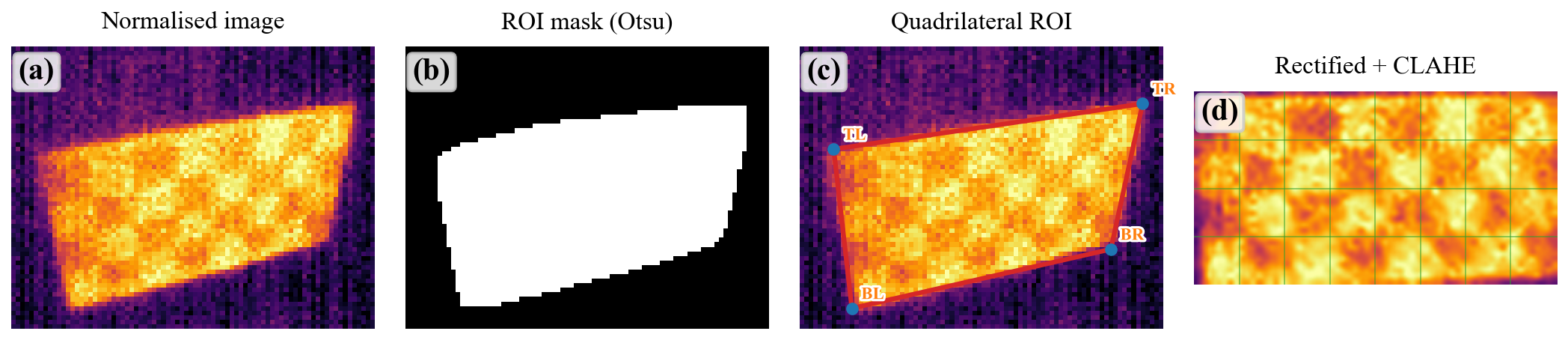}
\caption{ROI estimation and perspective rectification: (a)~normalised thermal image with percentile scaling; (b)~binary ROI mask obtained via Otsu thresholding and morphological refinement; (c)~quadrilateral enclosing the checkerboard derived from the ROI mask; (d)~image after rectification and histogram equalisation (CLAHE)~\cite{Zuiderveld1994} with the expected grid overlay.}
\label{fig:ir_roi_rectification}
\end{figure}

The ROI estimation and perspective rectification stages are illustrated in Figure~\ref{fig:ir_roi_rectification}.

\subsubsection{Hessian Saddle-Point Response}

Checkerboard corners correspond to saddle points of the image intensity function, at which the Hessian matrix exhibits eigenvalues of opposite sign. This saddle-point property of checkerboard corners has been exploited in prior work using the Hessian determinant~\cite{Beaudet1978,Placht2014}. For the rectified and upscaled image $I_{\mathrm{rect}}$, the Hessian matrix at each pixel is computed as
\begin{equation}
\mathcal{H}(x,y) = \begin{pmatrix} I_{xx} & I_{xy} \\ I_{xy} & I_{yy} \end{pmatrix},
\label{eq:hessian}
\end{equation}
where the second-order partial derivatives are obtained by applying Sobel operators to a Gaussian-smoothed version of the image with $\sigma = (p_x^* + p_y^*) / 8$. The saddle-point response map is defined as
\begin{equation}
S(x,y) = \max\!\bigl(-\det \mathcal{H}(x,y),\; 0\bigr) = \max\!\bigl(I_{xy}^2 - I_{xx}\,I_{yy},\; 0\bigr),
\label{eq:saddle_response}
\end{equation}
and normalised to $[0,1]$. Pixels with high values of $S$ correspond to locations where the intensity surface exhibits a saddle-point geometry characteristic of checkerboard corners. The saddle response defined in Eq.~\eqref{eq:saddle_response} constitutes the principal detection signal used by the Mean Shift stage.

\subsubsection{Mean Shift Corner Localisation}

Rather than applying global non-maximum suppression followed by grid assignment, the proposed algorithm initialises a separate Mean Shift \cite{Comaniciu2002} iteration from each nominal grid position. For an $(S_x - 1) \times (S_y - 1)$ grid of internal corners, the nominal position of corner $(r, c)$ in the rectified image is
\begin{equation}
\mathbf{g}_{r,c} = \bigl((c+1)\,p_x^*,\;\; (r+1)\,p_y^*\bigr).
\label{eq:nominal_grid}
\end{equation}

The nominal grid initialisation from Eq.~\eqref{eq:nominal_grid} ensures deterministic one-to-one assignment between grid nodes and detected corners. Mean Shift is applied to the response map $S$ using a rectangular window of size $\tfrac{1}{2}p_x^* \times \tfrac{1}{2}p_y^*$ centred at each $\mathbf{g}_{r,c}$. The algorithm iteratively shifts the window towards the local centre of mass of the response distribution until convergence \cite{Comaniciu2002}. A corner is accepted if the converged position $\mathbf{c}_{r,c}$ satisfies two conditions: the response value exceeds a minimum threshold ($S(\mathbf{c}_{r,c}) > 0.1$), and the displacement from the nominal position does not exceed the half-pitch in either axis. This strategy provides a one-to-one assignment between detected corners and grid nodes without requiring a separate matching stage.

\begin{figure}[H]
\centering
\includegraphics[width=0.95\linewidth]{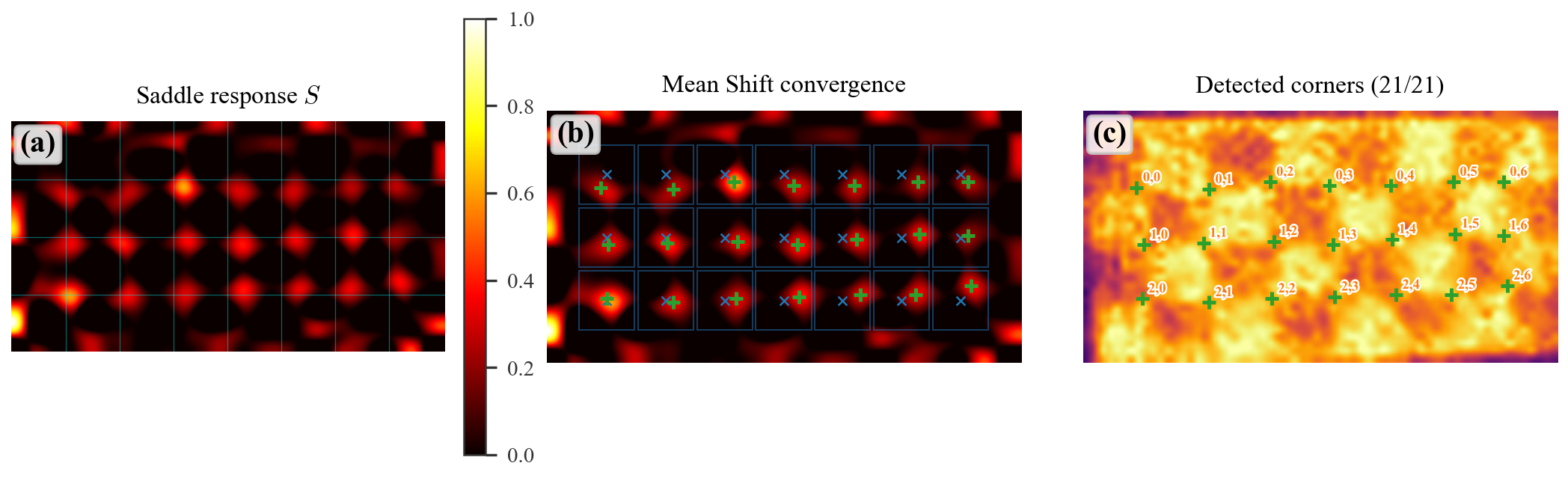}
\caption{Saddle-point detection and corner localisation: (a)~Hessian saddle response map $S$ with the expected grid overlay; (b)~Mean Shift convergence from nominal grid positions (blue crosses) to detected corner locations (green markers), with search windows indicated; (c)~detected corners overlaid in the rectified thermal image with grid indices.}
\label{fig:ir_saddle_meanshift}
\end{figure}

The Hessian saddle-point response and Mean Shift localisation results are illustrated in Figure~\ref{fig:ir_saddle_meanshift}.

\subsubsection{Quality Gate}
\label{sec:quality_gate}

After corner localisation, a frame-level quality gate evaluates whether the detected point set forms a geometrically consistent grid. The gate comprises two independent tests; a frame is accepted only if both are passed.

\paragraph{Test~1: Grid cell area regularity.} For each L-shaped triple of adjacent grid nodes $(r,c)$, $(r,c{+}1)$, $(r{+}1,c)$ that are all detected, the triangle area is computed using the generalised cross-product formula:
\begin{equation}
\Delta(\mathbf{u}, \mathbf{v}) = \tfrac{1}{2}\sqrt{\|\mathbf{u}\|^2 \|\mathbf{v}\|^2 - (\mathbf{u}^\top \mathbf{v})^2},
\label{eq:triangle_area}
\end{equation}
where $\mathbf{u} = \mathbf{p}(r,c{+}1) - \mathbf{p}(r,c)$ and $\mathbf{v} = \mathbf{p}(r{+}1,c) - \mathbf{p}(r,c)$. For a regular grid, all triangle areas should be close to the expected value $\Delta_{\mathrm{exp}} = \tfrac{1}{2}\,p_x^* \cdot p_y^*$. Let $\mathcal{A}$ denote the set of all valid L-triples; the mean and standard deviation of the collected areas are $\bar{\Delta} = |\mathcal{A}|^{-1}\sum_{i\in\mathcal{A}}\Delta_i$ and $\sigma_\Delta = \bigl(|\mathcal{A}|^{-1}\sum_{i\in\mathcal{A}}(\Delta_i - \bar{\Delta})^2\bigr)^{1/2}$, respectively. The coefficient of variation is defined as $\mathrm{CV}(\Delta) = \sigma_\Delta / \bar{\Delta}$. The gate requires $\mathrm{CV}(\Delta) < 0.15$ and a area ratio $\bar{\Delta} / \Delta_{\mathrm{exp}} \in (0.4,\; 2.5)$.

\paragraph{Test~2: Adjacent missing nodes.} Individual undetected grid nodes are tolerable, but two adjacent missing nodes (horizontally or vertically) indicate a contiguous gap in the grid that would compromise the local geometric constraint. Let $\mathcal{M}$ denote the set of missing grid nodes; the gate rejects the frame if
\begin{equation}
\exists\,(r,c) \in \mathcal{M} :\; (r,\,c{+}1) \in \mathcal{M} \;\vee\; (r{+}1,\,c) \in \mathcal{M}.
\label{eq:adjacent_missing}
\end{equation}

\begin{figure}[H]
\centering
\includegraphics[width=0.95\linewidth]{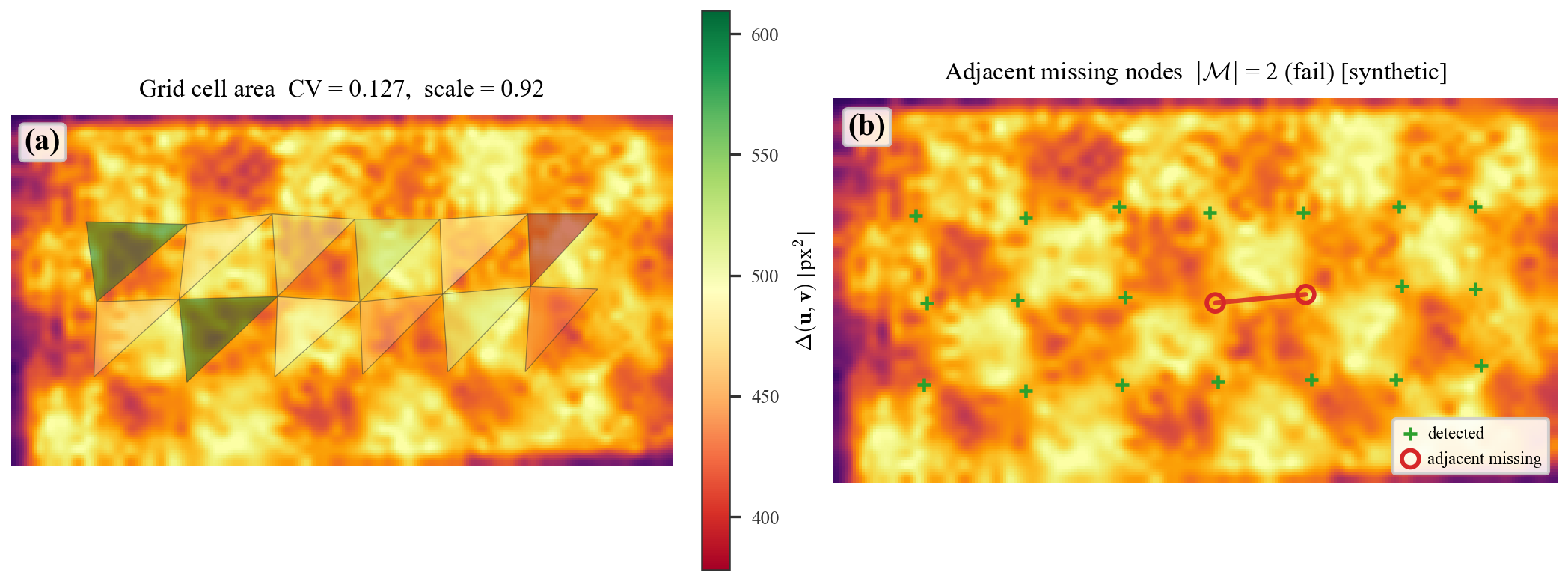}
\caption{Quality gate diagnostics for a representative frame: (a) triangle areas of adjacent grid cells colour-coded by deviation from the expected value; (b) adjacent missing node test — detected corners shown as green markers, with two synthetically removed adjacent nodes (red circles) connected by a red segment to illustrate a gate-failing configuration ($|\mathcal{M}|=2$).}
\label{fig:ir_quality_gate}
\end{figure}

Equations~\eqref{eq:triangle_area} and~\eqref{eq:adjacent_missing} jointly define the frame-level geometric consistency criterion used to reject unstable detections. A visualisation of the quality gate diagnostics is presented in Figure~\ref{fig:ir_quality_gate}.

\subsubsection{Perspective Projection to Original Coordinates}

\begin{figure}[H]
\centering
\includegraphics[width=0.85\linewidth]{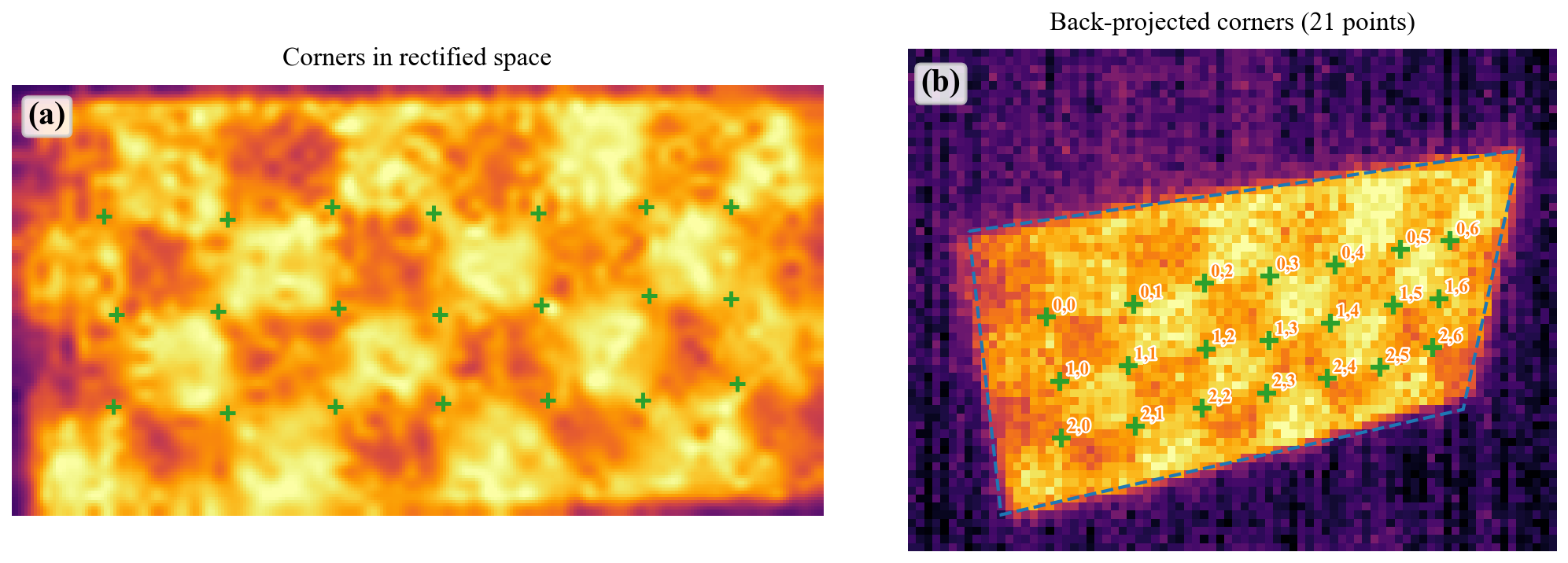}
\caption{Corner positions after perspective projection to the original thermal image coordinate system: (a)~corners in the rectified space; (b)~corresponding positions after inverse homography, overlaid on the normalised thermal image with grid labels.}
\label{fig:ir_warpback}
\end{figure}

Corners detected in the rectified and upscaled image are mapped back to the original image coordinate system by first rescaling by $1/\kappa$ to undo the upsampling, and then applying the inverse homography $H^{-1}$:
\begin{equation}
\mathbf{p}^{\mathrm{orig}} = H^{-1}\,\mathbf{p}^{\mathrm{rect}},
\label{eq:warp_back}
\end{equation}
where the transformation is applied in homogeneous coordinates. The resulting corner positions in the original image space, together with their grid indices, are directly compatible with the subsequent intrinsic calibration and stereo calibration stages. The back-projection defined by Eq.~\eqref{eq:warp_back} restores compatibility with the original thermal image coordinate system required by the calibration pipeline. An example of corners after perspective projection overlaid on the original thermal image is shown in Figure~\ref{fig:ir_warpback}.

\begin{figure}[H]
\centering
\includegraphics[width=\linewidth]{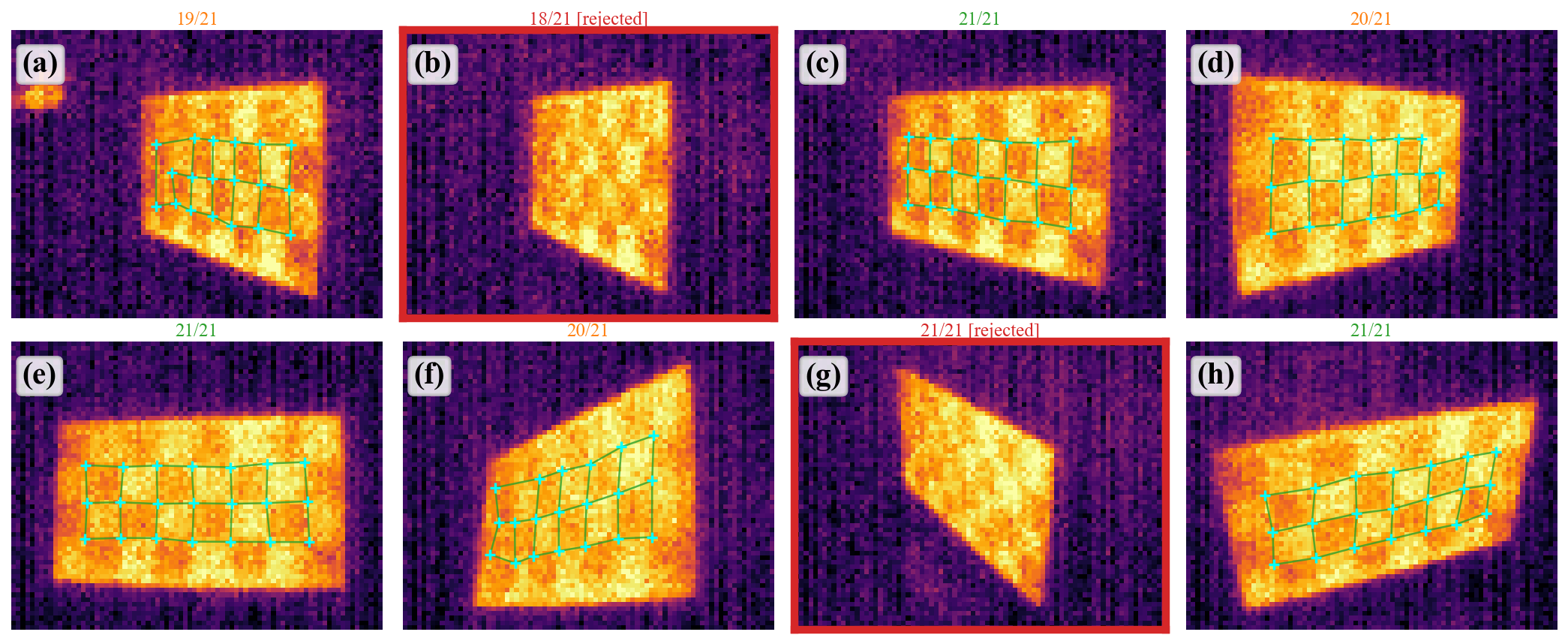}
\caption{Examples of checkerboard corner detection results (a--h) obtained by the proposed thermal corner detection algorithm on images of $80 \times 62$~px resolution. Detected corners are connected by grid lines; frames rejected by the quality gate are indicated by a red border. Detection statistics and gate diagnostics are annotated for each frame.}
\label{fig:ir_detection_results}
\end{figure}

Representative detection results obtained by the proposed algorithm in thermal images from a calibration session are presented in Figure~\ref{fig:ir_detection_results}. Despite the very low spatial resolution and varying thermal contrast across frames, the algorithm reliably detects a sufficient number of geometrically consistent corners for calibration, while the quality gate automatically rejects frames with degenerate or incomplete grid configurations.

\subsection{Frame Pairing Across Modalities}

During calibration sessions, the number of frames for which calibration points can be successfully detected may differ between the RGB and TIR modalities. This discrepancy arises from the distinct characteristics of the two cameras and from the higher susceptibility of thermal images to detection failures. However, stereo calibration requires unambiguously paired observations from both cameras that correspond to the same spatial configuration of the calibration object.

In the proposed procedure, frame pairing is performed based on the frame index within the acquisition sequence. For each thermal frame, the corresponding RGB frame with the same index is selected, while frames for which valid detection is not achieved in either modality are discarded. This strategy preserves geometric consistency between observations without requiring hardware-based temporal synchronisation between the cameras.

The use of index-based frame pairing is justified by the deterministic acquisition protocol and the absence of motion of the camera rig during image capture. Under these conditions, successive frames correspond to the same calibration object pose with respect to both cameras, even when image acquisition is performed sequentially rather than simultaneously.

The frame pairing procedure results in a set of corresponding RGB and TIR observations that can be directly utilised in subsequent stages of stereo calibration. At the same time, it enables the removal of inconsistent or incomplete data, thereby improving the overall stability and robustness of the calibration process.

\subsection{Harmonisation of RGB and TIR Point Sets}\label{sec:harmonisation}

Despite frame pairing across modalities, the sets of detected calibration points in RGB and thermal images differ both in cardinality and geometric structure. This discrepancy arises from the use of different calibration patterns and from the substantial resolution mismatch between the cameras. To enable correct stereo calibration, it is therefore necessary to harmonise the point representations across both modalities.

In RGB images calibrated using a ChArUco pattern, detection yields a dense grid of corner points with high localisation accuracy. In contrast, in thermal images, due to the low spatial resolution and the use of a classical checkerboard pattern, the number of detected points is significantly smaller. Direct use of these heterogeneous point sets would lead to inconsistencies in subsequent calibration stages.

In the proposed procedure, spatial subsampling of the RGB point set is applied in order to match the resolution and geometric layout of the points available in the thermal modality. Subsampling is performed by selecting every second corner in both the horizontal and vertical directions, thereby reducing the number of RGB points and forming a grid consistent with the structure of points detected in the thermal image.

After subsampling, a one-to-one correspondence between RGB and TIR points is ensured, together with consistent indexing of matching observations. The resulting point sets, along with their corresponding object points, can then be directly used in the stereo calibration process.

The harmonisation of point sets enables stable calibration of the RGB–TIR system despite the large resolution disparity and the use of different calibration patterns. This step forms the interface between the point detection procedure and the subsequent estimation of the geometric parameters of the system.

Let
\begin{equation}
\mathcal{P}_{\mathrm{RGB}} =
\left\{
\mathbf{p}_{\mathrm{RGB}}^{(k)} =
\begin{bmatrix}
x_{\mathrm{RGB}}^{(k)} \\
y_{\mathrm{RGB}}^{(k)}
\end{bmatrix}
\right\}_{k=1}^{N_{\mathrm{RGB}}}
\label{eq:p_rgb}
\end{equation}
denote the set of calibration points detected in the RGB image, and
\begin{equation}
\mathcal{P}_{\mathrm{TIR}} =
\left\{
\mathbf{p}_{\mathrm{TIR}}^{(m)} =
\begin{bmatrix}
x_{\mathrm{TIR}}^{(m)} \\
y_{\mathrm{TIR}}^{(m)}
\end{bmatrix}
\right\}_{m=1}^{N_{\mathrm{TIR}}}
\label{eq:p_tir}
\end{equation}
the set of points detected in the thermal image. The point sets defined in Eqs.~\eqref{eq:p_rgb}--\eqref{eq:p_tir} constitute the initial modality-specific observations prior to harmonisation.

Due to the higher density of the RGB point grid, the set $\mathcal{P}_{\mathrm{RGB}}$
is subjected to spatial subsampling. For each RGB point satisfying the relations
\begin{equation}
x_{\mathrm{RGB}} = 2\,x_{\mathrm{TIR}} + 1,
\qquad
y_{\mathrm{RGB}} = 2\,y_{\mathrm{TIR}} + 1,
\label{eq:subsampling}
\end{equation}
a unique corresponding position in the thermal point grid is defined. Equation~\eqref{eq:subsampling} defines the deterministic geometric correspondence between the dense RGB grid and the sparse thermal grid.

This transformation defines a subsampling operator that reduces the RGB point set
to a subset that is topologically consistent with the structure of points detected
in the thermal image. As a result, the obtained point sets exhibit a one-to-one correspondence
and consistent indexing, enabling their direct use in subsequent stages of stereo calibration.

\begin{figure}[H]
\centering

\begin{minipage}[t]{0.48\linewidth}
    \centering
    \includegraphics[width=\linewidth]{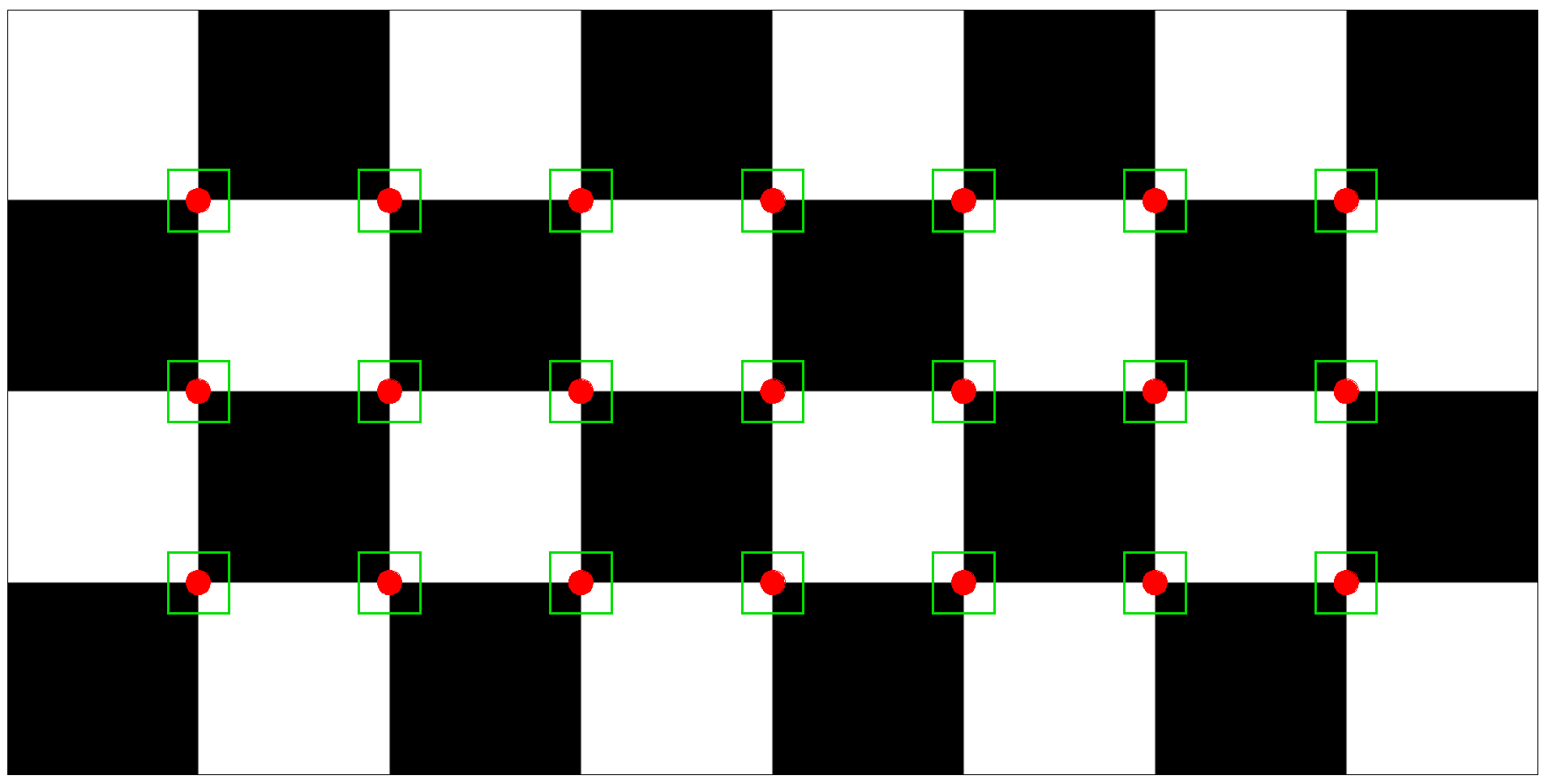}
\end{minipage}\hfill
\begin{minipage}[t]{0.48\linewidth}
    \centering
    \includegraphics[width=\linewidth]{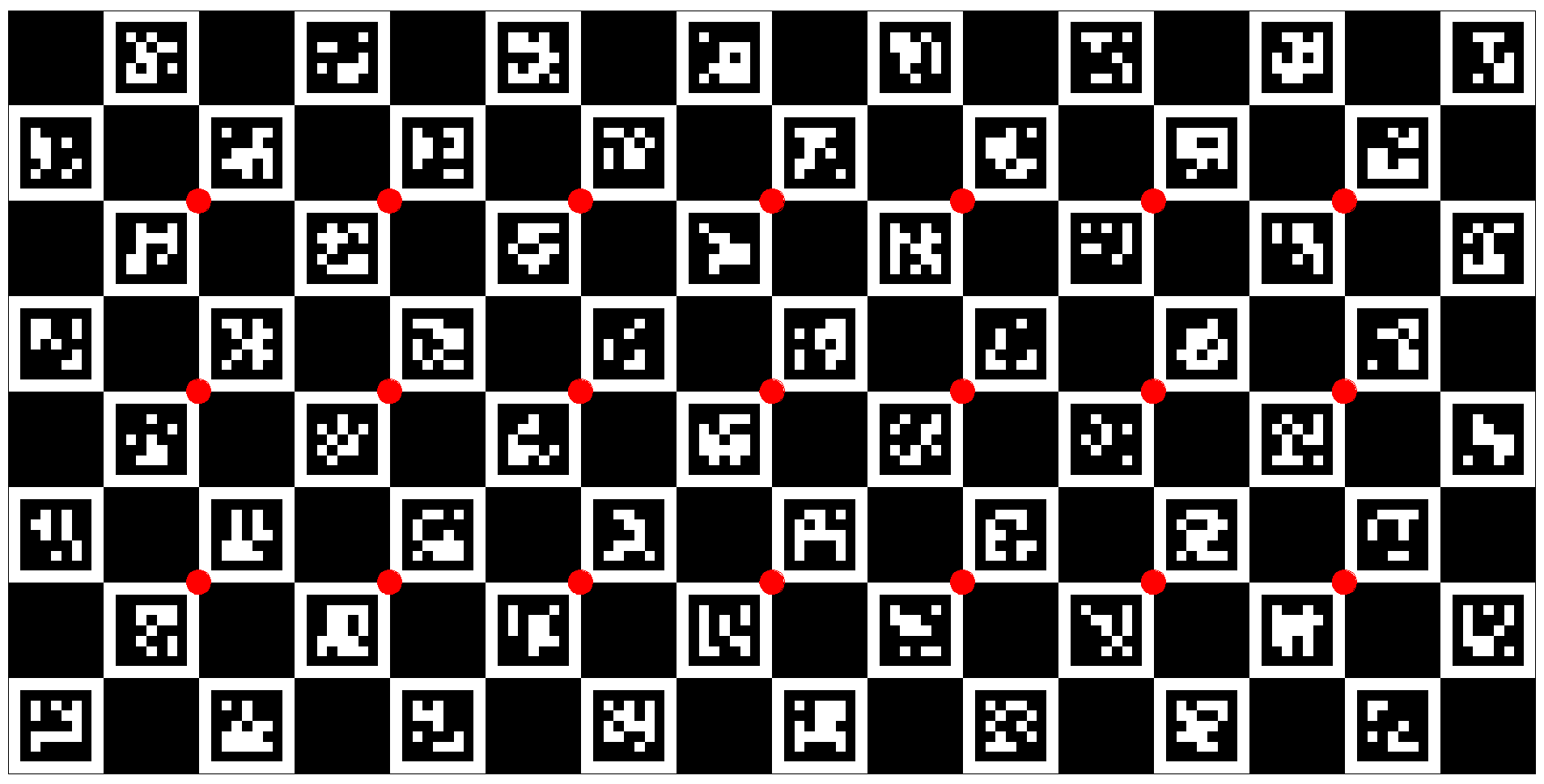}
\end{minipage}

\vspace{6pt}

\begin{minipage}[t]{0.48\linewidth}
    \centering
    \includegraphics[width=\linewidth]{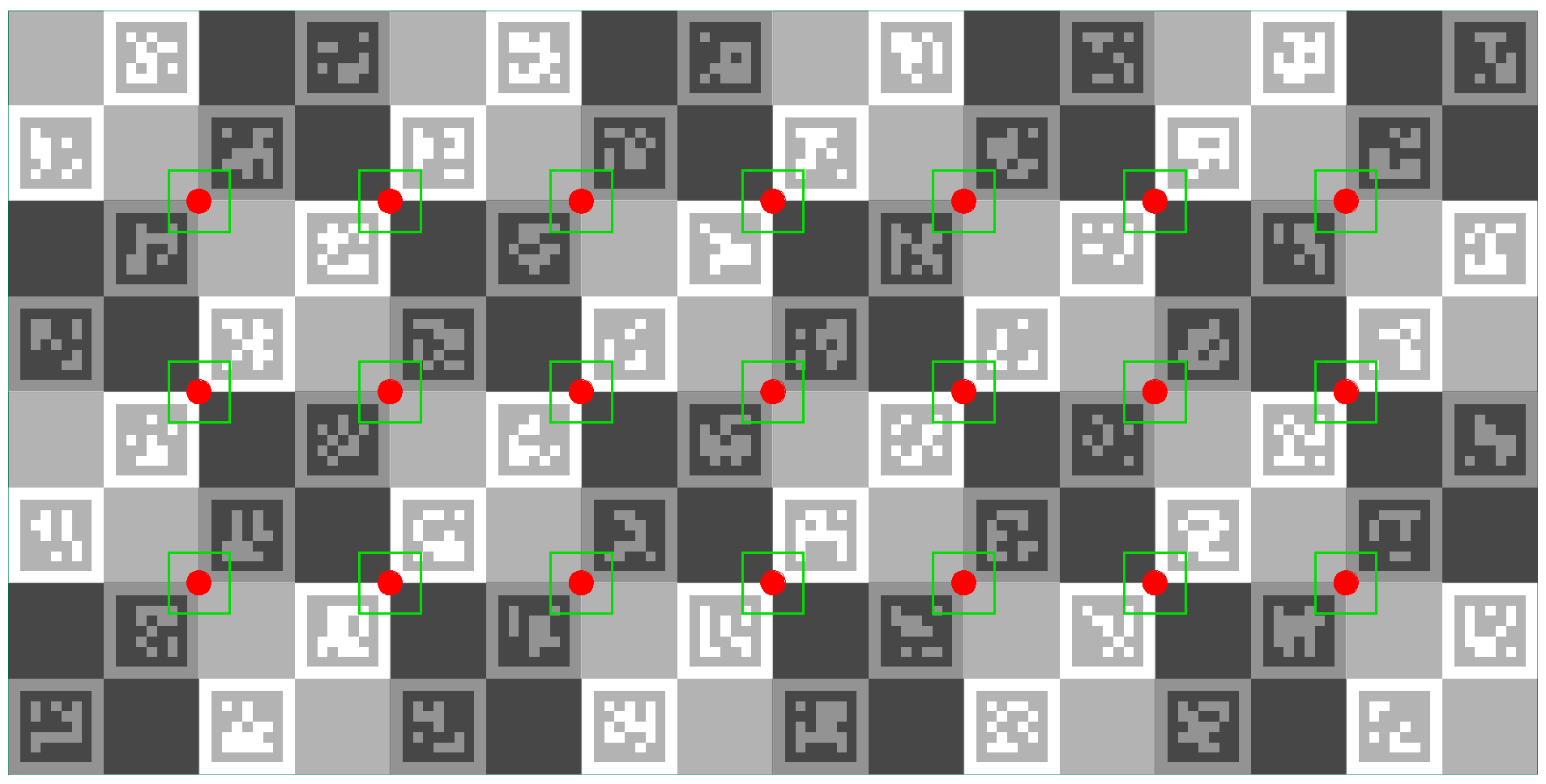}
\end{minipage}

\caption{Examples of calibration point sets detected in the RGB and TIR modalities for selected frames.}
\label{fig:points_unification_examples}
\end{figure}

Examples of calibration point sets detected in the RGB and TIR modalities for selected frames are shown in Figure~\ref{fig:points_unification_examples}.
The figure illustrates the substantial difference in point density between the high-resolution RGB images and the low-resolution thermal images prior to harmonisation.

\begin{figure}[H]
\centering
\includegraphics[width=\linewidth]{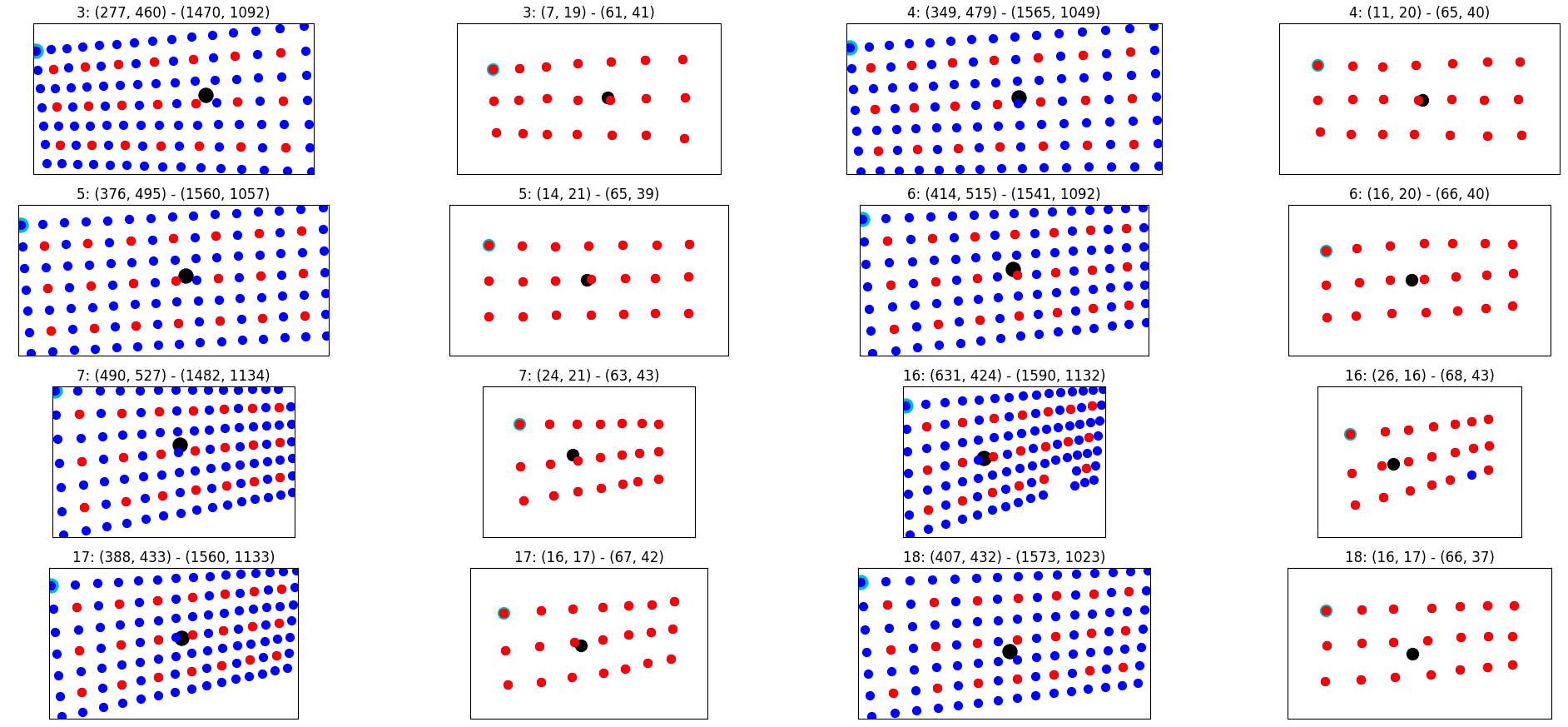}
\caption{Illustration of RGB point subsampling for harmonisation with the thermal point grid. The titles above each subplot indicate the frame ID and the bounding box of the corner region in the format $ID\colon (x_{\text{min}}, y_{\text{min}}) - (x_{\text{max}}, y_{\text{max}})$, where $(x_{\text{min}}, y_{\text{min}})$ represents the top-left corner and $(x_{\text{max}}, y_{\text{max}})$ represents the bottom-right corner. The black circle indicates the centre point of the image; only the calibration points within the bounding box are included in the figure.}
\label{fig:points_unification_subsampling}
\end{figure}

The effect of spatial subsampling applied to the RGB point set in order to achieve topological consistency with the thermal point grid is illustrated in Figure~\ref{fig:points_unification_subsampling}.
Despite the pronounced differences in image resolution and the number of detected points, the proposed subsampling strategy preserves one-to-one point correspondence and consistent indexing, enabling direct use of the harmonised point sets in subsequent stages of stereo calibration.

\section{RGB--Thermal Stereo Calibration}

Stereo camera calibration seeks to recover the rigid transformation between two camera coordinate frames — the rotation matrix $R \in \mathrm{SO}(3)$ and the translation vector $T \in \mathbb{R}^3$ — along with the intrinsic parameters of each sensor~\cite{Faugeras1993,HartleyZisserman2003}. This section describes the full calibration pipeline applied to the RGB--TIR pair. Section~\ref{sec:intrinsics} covers independent intrinsic calibration of each camera using the point correspondences obtained in Section~\ref{sec:detection_matching}. Section~\ref{sec:stereo_method} presents the standard stereo extrinsic estimation, and Section~\ref{sec:degeneracy} discusses the co-planar mounting geometry and the motivation for constraining the solution. Finally, Section~\ref{sec:constrained_ba} introduces the baseline-constrained bundle adjustment that exploits this geometry to improve calibration accuracy at the extreme resolution asymmetry ($80 \times 62$ vs.\ $2028 \times 1520$~px) of the sensor pair.

\subsection{Intrinsic Camera Calibration}
\label{sec:intrinsics}

Intrinsic calibration of the RGB and TIR cameras was performed independently for each modality using the standard pinhole camera model and classical calibration algorithms available in the OpenCV library \cite{OpenCV_Calib3d}. The objective of this stage was to estimate the intrinsic parameters, including the camera matrix and lens distortion coefficients, for each sensor prior to stereo calibration.

For the RGB camera, intrinsic calibration was carried out using images of the ChArUco calibration pattern acquired during data collection sessions. The use of the ChArUco pattern enabled unambiguous corner identification and precise sub-pixel localisation of image points, resulting in stable and repeatable estimation of intrinsic parameters across independent acquisition sessions.

Intrinsic calibration of the thermal (TIR) camera was performed using images of a classical checkerboard pattern displayed on the OLED screen. Despite the very low spatial resolution of the thermal sensor, the use of the active OLED screen combined with the dedicated corner detection algorithm described in Section~\ref{sec:ir_detector} allowed a sufficient number of valid corner detections to be obtained for reliable intrinsic parameter estimation.

Calibration was repeated across Sessions~III and~IV, conducted on different dates and under varying ambient conditions. The results, summarised in Table~\ref{tab:intrinsics}, demonstrate high consistency of the estimated intrinsic parameters across sessions. For the RGB camera, the focal length variation between sessions was below $0.1\%$, confirming the temporal stability of the optical system. For the TIR camera, the focal length estimates were consistent across both sessions, with $f_x \approx 107$~px and $f_y \approx 110$~px, both within the range expected from the sensor specification and field-of-view geometry.

The TIR sensor exhibits a non-square pixel geometry ($f_x \neq f_y$), which is accounted for in the stereo calibration model (Section~\ref{sec:stereo_method}).

The intrinsic calibration parameters obtained for both cameras serve as fixed inputs to the subsequent stereo extrinsic calibration stage.

\begin{table}[H]
\centering
\caption{Intrinsic calibration parameters of the RGB and TIR cameras.
Sessions~III and~IV correspond to the finalised acquisition protocol; Session~I is included to demonstrate long-term intrinsic stability of the RGB camera. Distortion coefficients $k_3$, $p_1$, and $p_2$ were constrained to zero for the TIR camera (see Section~\ref{sec:stereo_method}).}
\label{tab:intrinsics}
\begin{tabular}{llccccccc}
\hline
Camera & Session & Frames & $f_x$ [px] & $f_y$ [px] & $c_x$ [px]
  & $c_y$ [px] & $k_1$ / $k_2$ & RMS [px] \\
\hline
RGB & I   & 44 & 1939.07 & 1939.07 & 969.70 & 781.56
  & --- & 0.035 \\
RGB & III & 23 & 1940.65 & 1940.65 & 966.86 & 784.26
  & --- & 0.021 \\
RGB & IV  & 43 & 1939.18 & 1939.18 & 971.55 & 784.79
  & --- & 0.030 \\
\hline
TIR & III & 23 & 106.81 & 109.75 & 40.82 & 29.34
  & $-0.51$ / $-8.60$ & 0.154 \\
TIR & IV  & 38 & 107.56 & 109.81 & 42.60 & 35.75
  & $-0.11$ / $-0.01$ & 0.093 \\
\hline
\end{tabular}
\end{table}

The focal length and principal point estimates for the TIR camera are consistent across both sessions. However, the second radial distortion coefficient $k_2$ exhibits large variation ($-8.60$ in Session~III vs $-0.01$ in Session~IV). The anomalous value in Session~III is attributed to the smaller number of accepted frames and less favourable thermal contrast conditions, which reduce the observability of higher-order distortion parameters at $80 \times 62$~px. This instability further motivates the decision to constrain $k_3 = 0$ and confirms that only low-order distortion coefficients can be reliably estimated at this resolution. The stereo calibration reported in the following sections uses the intrinsic parameters from Session~IV.

\subsection{Stereo Extrinsic Calibration}
\label{sec:stereo_method}

The objective of RGB--TIR stereo calibration is to estimate the rotation matrix $R$ and translation vector $T$ that describe the rigid transformation between the coordinate systems of the RGB and thermal cameras. Since the two sensors are mounted on a rigid bracket with a known physical geometry, the expected stereo translation is predominantly horizontal, with $T \approx [T_x,\; 0,\; 0]^\top$ and $T_x \approx 0.03$~m corresponding to the inter-camera baseline.

Stereo calibration was performed using the \texttt{stereoCalibrateExtended} function from the OpenCV library \cite{OpenCV_Calib3d}, with intrinsic parameters of both cameras held fixed at their independently estimated values (\texttt{CALIB\_FIX\_INTRINSIC}). Paired observations of calibration object points, detected independently in both modalities, were matched using the algebraic grid correspondence described in Section~\ref{sec:harmonisation}. Views with fewer than eight matched point pairs were discarded. In Session~IV, the RGB detection pipeline accepted 43~frames and the TIR quality gate (Section~\ref{sec:quality_gate}) passed 38~frames. After index-based frame pairing and rejection of views with fewer than eight matched point pairs, 36~matched views remained for stereo calibration.

An important aspect of the intrinsic model adopted for the TIR camera concerns the pixel geometry of the sensor. According to the manufacturer specification, the TIR camera has a nominal field of view of $45^{\circ} \times 45^{\circ}$ across a non-square pixel grid of $80 \times 62$~px. Because the same angular range is sampled by a different number of pixels along each axis, the effective focal lengths in the horizontal and vertical directions are not equal. Consequently, the constraint $f_x = f_y$ (\texttt{CALIB\_FIX\_ASPECT\_RATIO}) was not enforced during intrinsic calibration, allowing the optimisation to recover the true pixel geometry from the calibration data. The calibrated focal lengths ($f_x \approx 107$~px, $f_y \approx 110$~px; Table~\ref{tab:intrinsics}) deviate from the values predicted by the nominal $45^{\circ}$~FOV specification, which is not uncommon for low-cost thermal sensors whose effective optical parameters may differ from the datasheet values. In addition, to prevent overfitting on the very low-resolution thermal images, the third radial distortion coefficient $k_3$ was fixed to zero (\texttt{CALIB\_FIX\_K3}), as its inclusion led to numerically degenerate solutions in unconstrained optimisation. Tangential distortion coefficients $p_1$ and $p_2$ were likewise fixed to zero (\texttt{CALIB\_ZERO\_TANGENT\_DIST}), since these parameters cannot be reliably separated from sensor misalignment effects at a resolution of $80 \times 62$ pixels.

Using the standard unconstrained stereo calibration procedure on 36 matched views from Session~IV, the following extrinsic parameters were obtained: $T_x = 3.46$~cm, $T_y = 0.41$~cm, and $T_z = 3.48$~cm, yielding a total baseline of $\|T\| = 4.93$~cm. The overall root mean square (RMS) reprojection error was 0.453~px, with per-modality values of 0.173~px for the RGB camera and 0.614~px for the TIR camera. The per-modality RMS values were computed by aggregating the per-view reprojection errors returned by \texttt{stereoCalibrateExtended}, weighted by the number of point correspondences in each view. Specifically, for each modality $m \in \{\mathrm{RGB}, \mathrm{TIR}\}$:
\begin{equation}
\mathrm{RMS}_m = \sqrt{\frac{\sum_{i=1}^{N} e_{m,i}^2 \cdot n_i}{\sum_{i=1}^{N} n_i}},
\label{eq:rms_modality}
\end{equation}
where $e_{m,i}$ is the per-view reprojection error for modality $m$ in view $i$, and $n_i$ is the number of matched point pairs in that view. The overall RMS reported in Table~\ref{tab:stereo_comparison} is the value returned directly by \texttt{stereoCalibrateExtended}, computed as the root mean square of all individual reprojection residuals pooled across both modalities and all views:
\begin{equation}
\mathrm{RMS}_{\mathrm{total}} = \sqrt{\frac{1}{2\sum_{i=1}^{N} n_i} \sum_{i=1}^{N} \sum_{m} \sum_{j=1}^{n_i} \left\| \mathbf{p}_{m,ij} - \hat{\mathbf{p}}_{m,ij} \right\|^2},
\label{eq:rms_total}
\end{equation}
where $\mathbf{p}_{m,ij}$ and $\hat{\mathbf{p}}_{m,ij}$ denote the observed and reprojected image coordinates, respectively. While the reprojection residuals are low, the estimated translation vector is physically inconsistent with the known rig geometry: the $T_z$ component, which should be approximately zero for co-planar cameras, accounts for a substantial portion of the total baseline. This systematic discrepancy motivated a detailed investigation of its geometric origin, presented in the following subsection.

\subsection{Co-planar Sensor Mounting and Constraint Motivation}
\label{sec:degeneracy}

The RGB and TIR sensors are mounted in a co-planar configuration: both optical axes are nearly perpendicular to the same mounting plane, so the depth component of the inter-sensor translation is nominally zero ($T_z = 0$). This physical geometry directly justifies fixing $T_z$ in the calibration.

In the unconstrained stereo calibration, the optimiser consistently returned $T_z \approx 3$--$4$~cm regardless of frame selection, a value with no physical justification given the known rig geometry. The improvement in TIR reprojection error was marginal ($0.74$~px at $T_z = 0$ versus $0.61$~px at $T_z = 3.48$~cm), indicating that the unconstrained estimate absorbed measurement noise rather than reflecting a true geometric offset. These observations motivated the constrained formulation described in the following subsection, in which $T_z = 0$ is enforced explicitly.

\subsection{Baseline-Constrained Bundle Adjustment}
\label{sec:constrained_ba}

To resolve the geometric degeneracy identified in the preceding
subsection, the stereo calibration problem was reformulated as a
bundle adjustment~(BA) with an explicit structural constraint on
the translation vector. Rather than allowing the optimiser to
estimate all three components of~$T$ freely, the depth
component~$T_z$ was fixed to zero, reducing the translational
degrees of freedom from three to two. This constraint is
justified by the known physical geometry of the camera rig, in
which the RGB and TIR sensors are mounted in a co-planar
configuration.

Imposing such geometric priors on the translation vector is
supported by the constrained optimisation framework established
in~\cite{Ma2004, HartleyZisserman2003}, and has been shown to
improve the conditioning of joint multi-sensor calibration
problems~\cite{Furgale2013}. In the present setup, the extremely
low TIR resolution combined with near-planar acquisition geometry
render the estimation of~$T_z$ numerically unstable without this
constraint, which further reinforces its application here.

\subsubsection{Problem Formulation}

The stereo extrinsic calibration is formulated as a nonlinear least-squares problem solved using the Levenberg--Marquardt algorithm \cite{Triggs2000}. The cost function minimises the sum of squared reprojection errors across both modalities and all calibration views:
\begin{equation}
\min_{R,\, T,\, \{R_i, t_i\}} \;
\sum_{i=1}^{N} \sum_{j=1}^{M_i}
\left[
\left\| \mathbf{p}_{\mathrm{RGB},ij} - \pi\!\left(K_{\mathrm{RGB}},\; R_i,\; t_i,\; P_j\right) \right\|^2
+
\left\| \mathbf{p}_{\mathrm{TIR},ij} - \pi\!\left(K_{\mathrm{TIR}},\; R \cdot R_i,\; R \cdot t_i + T,\; P_j\right) \right\|^2
\right],
\label{eq:ba_cost}
\end{equation}
where $\pi(\cdot)$ denotes the pinhole projection function, $K_{\mathrm{RGB}}$ and $K_{\mathrm{TIR}}$ are the intrinsic matrices of the respective cameras (held fixed), $P_j$ are the three-dimensional object points defined in the calibration object coordinate system, and $(R_i, t_i)$ is the pose of the calibration board in the RGB camera frame for view $i$. The stereo extrinsic parameters $(R, T)$ describe the rigid transformation from the RGB to the TIR camera coordinate system.

The key difference from the standard \texttt{stereoCalibrate} implementation lies in the parameterisation of the translation vector. In the unconstrained formulation, $T = [T_x,\; T_y,\; T_z]^\top$ comprises three free parameters. In the proposed constrained formulation,
\begin{equation}
T = [T_x,\; T_y,\; 0]^\top,
\label{eq:tz_constraint}
\end{equation}
reducing the translational degrees of freedom to two. The rotation $R$ is parameterised using the Rodrigues vector representation, yielding three rotational parameters.

The full optimisation state vector is
\begin{equation}
\mathbf{x} = \left[\, r_x,\; r_y,\; r_z,\; T_x,\; T_y,\;\;
\mathbf{r}_1^\top,\; \mathbf{t}_1^\top,\;\;
\ldots,\;\;
\mathbf{r}_N^\top,\; \mathbf{t}_N^\top
\,\right]^\top,
\label{eq:ba_state}
\end{equation}
comprising 5 stereo parameters and $6N$ per-view board pose parameters. For $N = 36$ calibration views, the total number of optimisation variables is $5 + 6 \times 36 = 221$. All intrinsic parameters remain fixed throughout the optimisation.

\subsubsection{Initialisation}

The optimisation was warm-started from the result of the unconstrained OpenCV \texttt{stereoCalibrate} solution, with $T_z$ set to zero. Per-view board poses were initialised by solving the Perspective-n-Point (PnP) problem independently for each RGB view using \texttt{cv2.solvePnP}. This two-stage initialisation strategy follows standard practice in calibration pipelines \cite{Triggs2000} and ensures convergence to a physically consistent local minimum.

\subsubsection{Results}

The constrained bundle adjustment converged in 18 iterations. A comparison of the stereo extrinsic parameters and reprojection errors obtained with the unconstrained OpenCV solution and the proposed constrained BA is presented in Table~\ref{tab:stereo_comparison}.

\begin{table}[H]
\centering
\caption{Comparison of stereo extrinsic calibration results: unconstrained OpenCV \texttt{stereoCalibrate} versus the proposed baseline-constrained bundle adjustment ($T_z = 0$). Both methods use the same set of 36 matched views from Session~IV.}
\label{tab:stereo_comparison}
\begin{tabular}{lcc}
\hline
Parameter & OpenCV (unconstrained) & Constrained BA ($T_z = 0$) \\
\hline
$T_x$ [cm]          & 3.46  & 3.27 \\
$T_y$ [cm]          & 0.41  & 0.04 \\
$T_z$ [cm]          & 3.48  & 0.00 \\
$\|T\|$ [cm]        & 4.93  & 3.27 \\
RMS total [px]       & 0.453 & 0.382 \\
RMS RGB [px]         & 0.173 & 0.175 \\
RMS TIR [px]         & 0.614 & 0.744 \\
Rotation angle       & $4.08^{\circ}$ & $3.88^{\circ}$ \\
\hline
\end{tabular}
\end{table}

The constrained solution yields a baseline of $\|T\| = 32.7$~mm, which is consistent with the mechanical mount separation of approximately 30~mm. The vertical translation component $T_y \approx 0$ confirms co-planar vertical alignment of the two sensors. The rotation angle of $3.88^{\circ}$ represents a small misalignment between the cameras, consistent with the tolerances of manual assembly.

The increase in TIR reprojection error from 0.614~px to 0.744~px ($+0.13$~px) represents the true cost of enforcing the $T_z = 0$ constraint and confirms that the $T_z = 3.48$~cm returned by the unconstrained optimisation was compensating for measurement noise rather than encoding a genuine geometric offset. Notably, the total RMS decreased from 0.453~px to 0.382~px, indicating that the constrained solution achieves a more balanced distribution of residuals across both modalities.

The parameter \texttt{fix\_tz} in the proposed formulation can be set to any known value, generalising the method to configurations in which the depth offset between sensors is non-zero but known from mechanical measurements.

\paragraph{\textit{Comparison with low-resolution stereo baselines}}
The TIR reprojection RMS of $0.744$~px reported in Table~\ref{tab:stereo_comparison} is higher than the quarter-pixel error reported by Zoetgnand\'e et al.~\cite{Zoetgnande2019} for a comparable $80 \times 60$~pixel configuration. This difference reflects two fundamental distinctions between the two settings. First, Zoetgnand\'e et al.\ calibrate a homogeneous thermal--thermal stereo pair in which both views share the same image characteristics and subpixel phase-congruency features, whereas the present system couples a thermal view with a $2028 \times 1520$~pixel RGB view whose order-of-magnitude higher localisation precision amplifies residuals attributable to the TIR camera. Second, the planar-calibration-object degeneracy addressed by the $T_z = 0$ constraint (Section~\ref{sec:degeneracy}) does not arise in thermal--thermal stereo, where both cameras observe the pattern at comparable resolutions. The passive RGB--TIR pipeline of Piccinelli et al.~\cite{Piccinelli2024} targets a heterogeneous configuration similar to ours but does not report quantitative reprojection metrics at an explicit resolution ratio, precluding a direct numerical comparison.

\subsubsection{Interpretation and Consistency}

The constrained bundle adjustment yielded a stereo baseline of $\|T\| = 32.7$~mm, which is in good agreement with the nominal inter-camera distance of approximately 30~mm. The residual difference of approximately 2.7~mm can be attributed to manufacturing tolerances of the 3D-printed mount and to the offset between the physical camera housing and its optical centre. The vertical component $T_y = 0.4$~mm confirms co-planar alignment, and the rotation angle of $3.88^{\circ}$ reflects a small misalignment consistent with manual assembly tolerances.

The per-modality reprojection errors reflect the expected asymmetry between the two sensors. The RGB camera achieves an RMS of 0.175~px, consistent with sub-pixel ChArUco detection at high resolution. The TIR camera exhibits an RMS of 0.744~px, reflecting the combined effects of low sensor resolution, elevated noise, and limited geometric precision at $80 \times 62$~px. Despite this asymmetry, the TIR reprojection error remains below one pixel, indicating geometric consistency with the estimated stereo model.

The increase in TIR reprojection error from 0.614~px (unconstrained) to 0.744~px ($+0.13$~px) represents the cost of enforcing the $T_z = 0$ constraint. This trade-off is characteristic of constrained estimation, in which prior geometric knowledge redistributes residuals from weakly observable parameters towards physically meaningful image-domain errors~\cite{Triggs2000, HartleyZisserman2003}. The consistency of the rotation estimates across all evaluated methods (inter-method differences below 0.005~rad) confirms that the rotational parameters are well-conditioned and robust to the choice of estimation algorithm~\cite{HartleyZisserman2003}.

\section{Applications and Discussion}

\subsection{Relevance to Building Energy Assessment}

The calibrated RGB--TIR stereo system enables consistent spatial alignment of thermal and geometric data, which constitutes a prerequisite for quantitative building envelope analysis. Thermally augmented 3D models derived from co-registered imagery allow the localisation of thermal bridges, insulation defects, and anomalous heat transfer paths within a coherent geometric framework. When combined with known boundary conditions, such models can support the estimation of effective thermal transmittance and provide input data for building energy simulations. While the present study focuses on the calibration methodology itself, the demonstrated projection quality (Sections~\ref{sec:charuco_fusion}--\ref{sec:depthpro_fusion}) indicates that the obtained stereo parameters are suitable for downstream building physics applications.

\subsection{Applications and Discussion: Thermal Projection onto RGB Using the ChArUco Board Pose}
\label{sec:charuco_fusion}

In order to practically verify the quality of the stereo calibration between the RGB and TIR cameras, an image fusion scenario based on a ChArUco board affixed to a window surface was employed. As a first step, both the RGB and TIR images were corrected for optical radial distortion using the calibrated intrinsic parameters and updated projection matrices, ensuring that subsequent geometric operations are consistent with the pinhole camera model.

In the RGB image, ChArUco markers were detected and corners were interpolated, after which the pose of the board was estimated using a Perspective-n-Point (PnP) method. This procedure yielded rotation and translation vectors expressed with respect to the RGB camera coordinate system. The visualisation of the coordinate axes overlaid on the RGB image (Fig.~\ref{fig:fusion_compare}(a)) provides a direct and intuitive validation of both the correctness of feature detection and the stability of the pose estimation.

From the estimated board pose, the geometric plane of the window surface was reconstructed in the RGB camera coordinate system. The unit normal vector~$\mathbf{n} = (n_x, n_y, n_z)^\top$ was obtained as the third column of the PnP rotation matrix (corresponding to the board's local $Z$-axis expressed in the RGB camera frame), and a scalar offset~$d = \mathbf{n}^\top \mathbf{c}$ was computed from the board centre~$\mathbf{c}$ (the PnP translation vector), yielding the plane equation:
\begin{equation}
    n_x X + n_y Y + n_z Z = d.
    \label{eq:plane}
\end{equation}

For each pixel $(u, v)$ of the undistorted RGB image, a viewing ray was formed by back-projecting through the updated intrinsic matrix $K_{\mathrm{RGB,new}}$, giving a direction vector $(x', y', 1)^\top$ in normalised camera coordinates, so that the corresponding 3D point at depth $Z$ is $\mathbf{X} = Z(x', y', 1)^\top$. Substituting this parametrisation into Eq.~\eqref{eq:plane} and solving for $Z$ yields the per-pixel depth of the ray–plane intersection:
\begin{equation}
    Z(u,v) = \frac{d}{\,n_x x' + n_y y' + n_z\,},
    \label{eq:dynamic_z}
\end{equation}
This formulation accounts for the fact that a planar surface observed at an oblique angle subtends a continuously varying depth across the image: pixels near the closer edge of the window receive a smaller $Z$, while pixels near the farther edge receive a larger $Z$. The 3D point reconstructed for each RGB pixel is accordingly $\mathbf{X}_{\mathrm{rgb}} = Z(u,v)\,(x', y', 1)^\top$.

These 3D points were subsequently transformed into the coordinate system of the TIR camera using the stereo calibration parameters — the rotation matrix~$R$ and the translation vector~$T$:
\begin{equation}
    \mathbf{X}_{\mathrm{ir}} = R\,\mathbf{X}_{\mathrm{rgb}} + T.
    \label{eq:stereo_rigid}
\end{equation}
After projection onto the TIR image plane via the TIR intrinsic matrix, per-pixel transformation maps were obtained and used to remap the thermal image onto the RGB image, resulting in a semi-transparent pseudocolour thermal overlay (Fig.~\ref{fig:fusion_compare}(b)).

The obtained results demonstrate a high level of geometric consistency between the RGB and TIR images in the region corresponding to the board plane, confirming the correctness of the stereo calibration. Because the depth is derived from the actual plane geometry rather than a single constant value, the method naturally compensates for the oblique viewing angle and avoids the systematic parallax shift that would result from a fixed $Z$ assumption. The inherent limitation is that the accuracy of the projection degrades for scene elements that deviate significantly from the calibration plane. Additionally, the visualisation of the TIR signal distribution in the form of isolines (Fig.~\ref{fig:fusion_compare}(c)) facilitates the interpretation of local thermal gradients in the context of the RGB scene structure.

\subsection{Applications and Discussion: Thermal-to-RGB Projection Using Depth Estimation with DepthPro}
\label{sec:depthpro_fusion}

\begin{figure}[H]
    \centering

    \begin{minipage}[t]{0.32\linewidth}
        \centering
        \includegraphics[width=\linewidth]{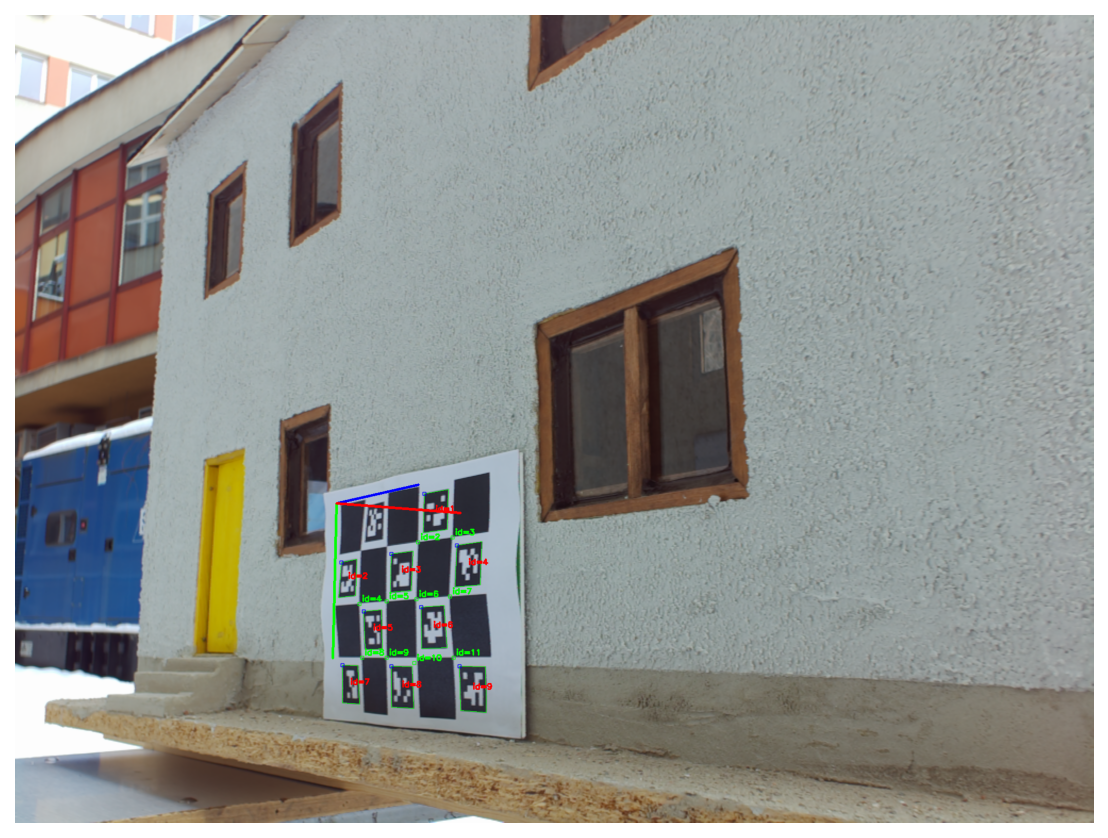}
        \small (a)
    \end{minipage}
    \hfill
    \begin{minipage}[t]{0.32\linewidth}
        \centering
        \includegraphics[width=\linewidth]{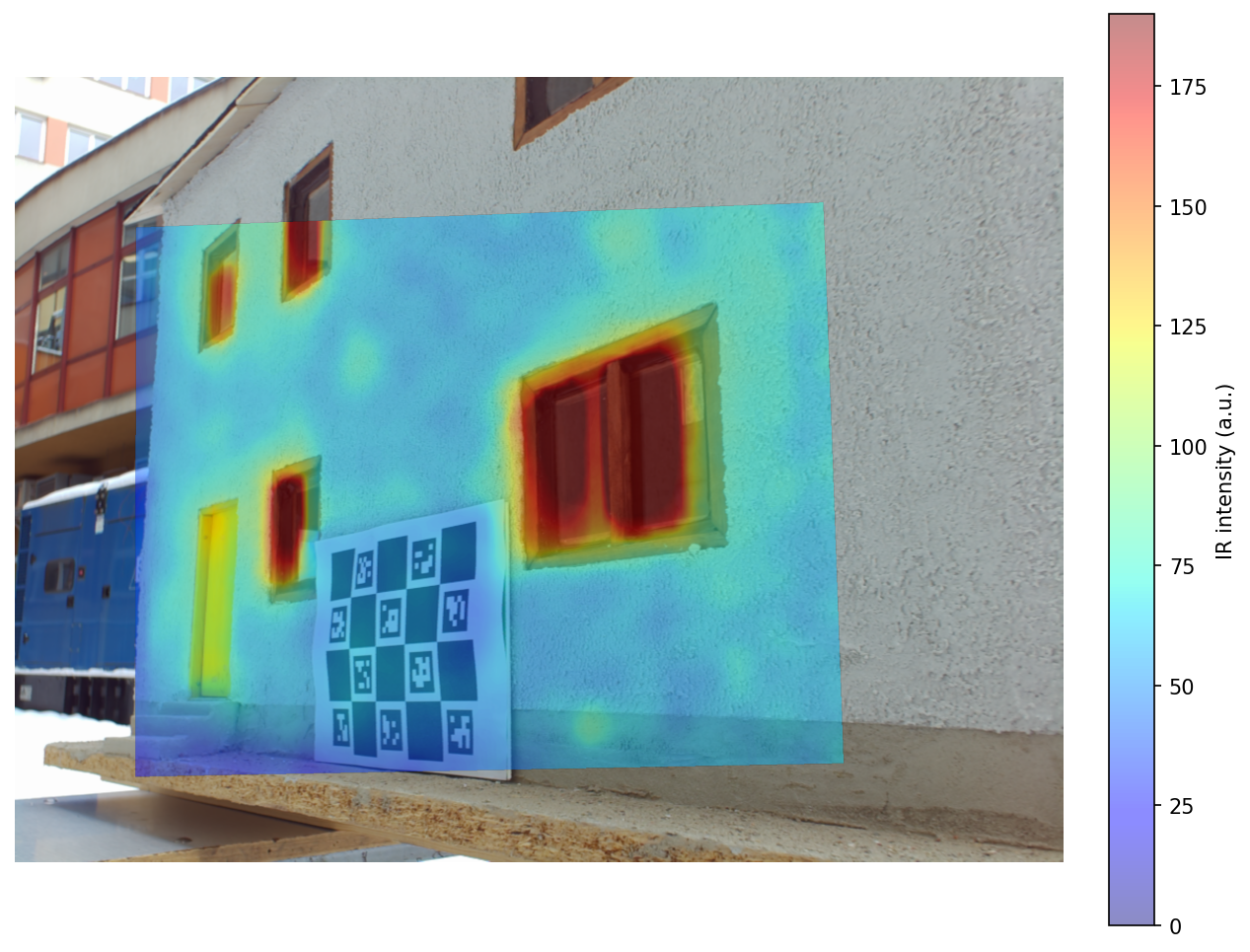}
        \small (b)
    \end{minipage}
    \hfill
    \begin{minipage}[t]{0.32\linewidth}
        \centering
        \includegraphics[width=\linewidth]{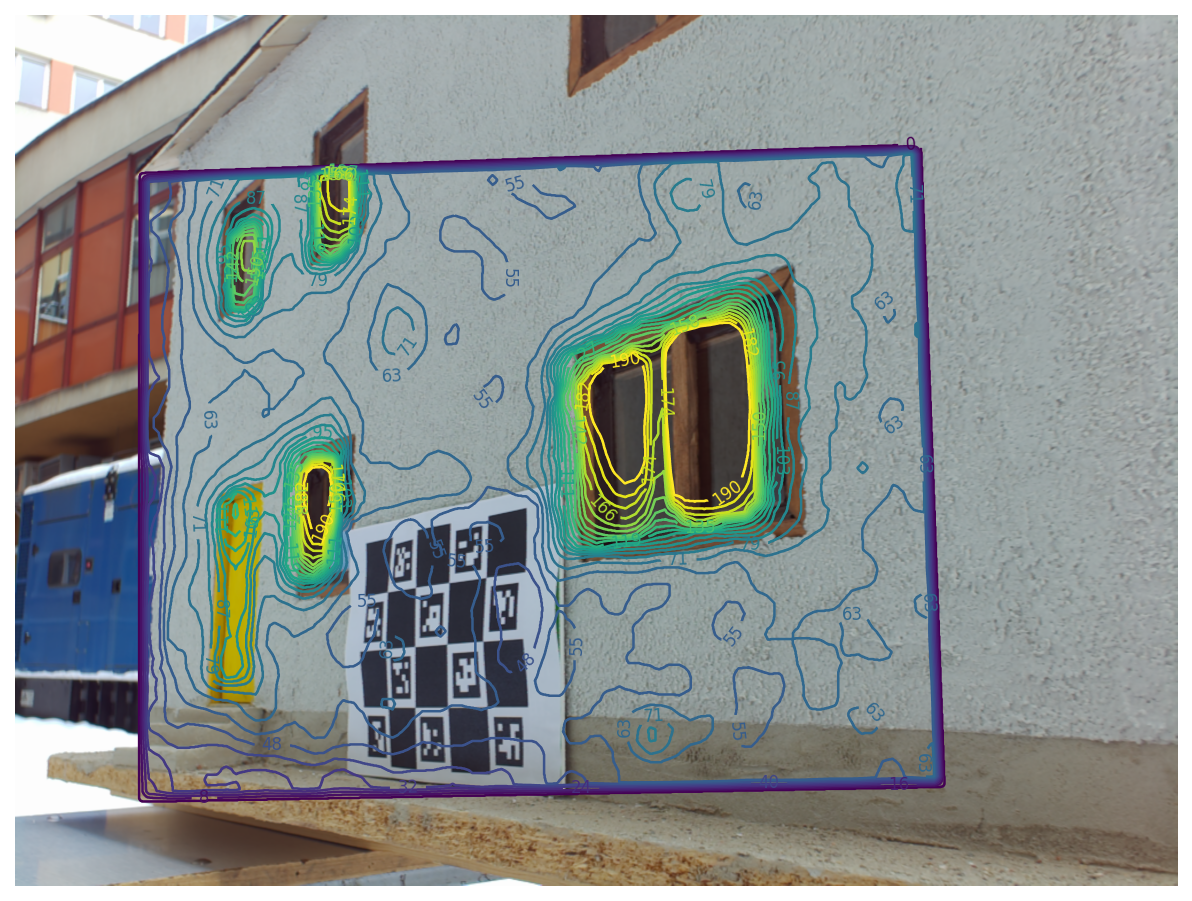}
        \small (c)
    \end{minipage}

    \vspace{0.5em}

    \begin{minipage}[t]{0.32\linewidth}
        \centering
        \includegraphics[width=\linewidth]{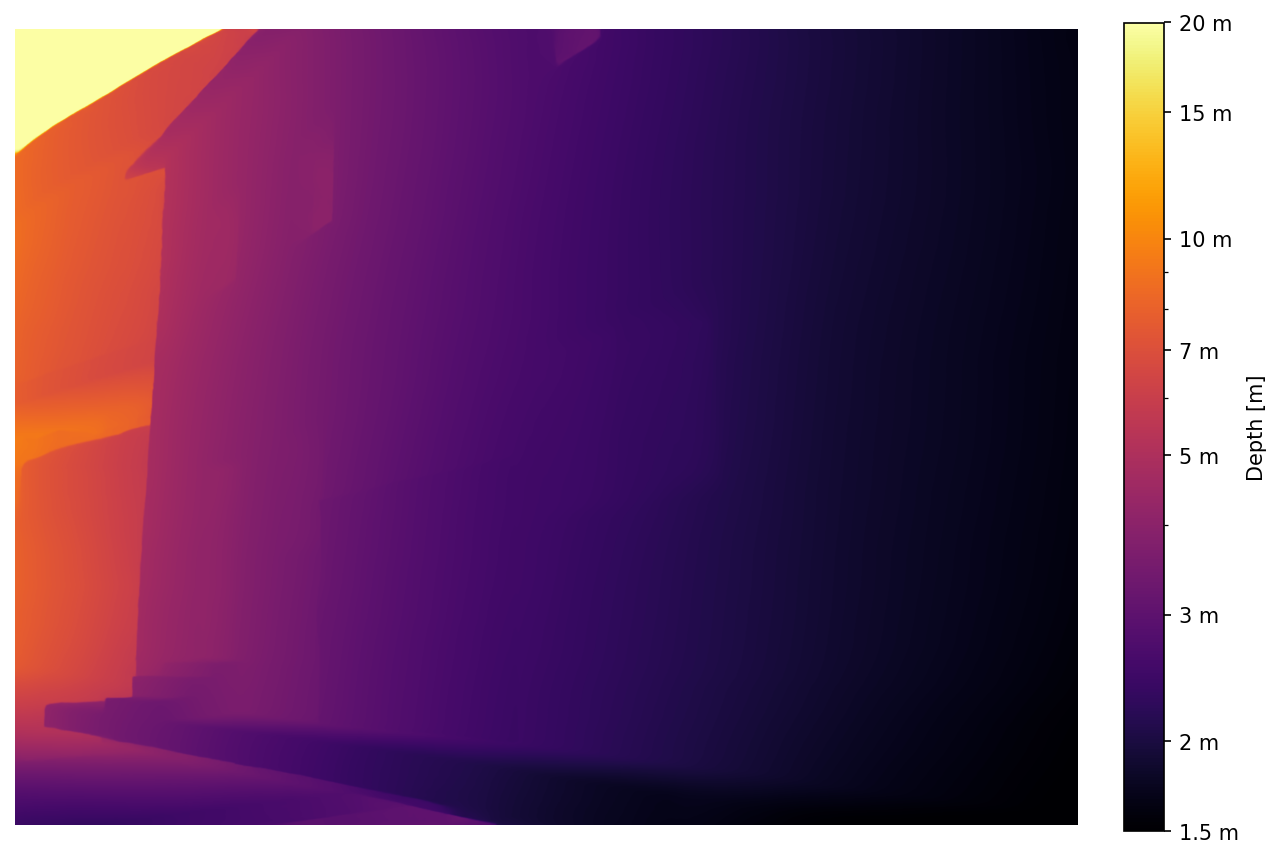}
        \small (d)
    \end{minipage}
    \hfill
    \begin{minipage}[t]{0.32\linewidth}
        \centering
        \includegraphics[width=\linewidth]{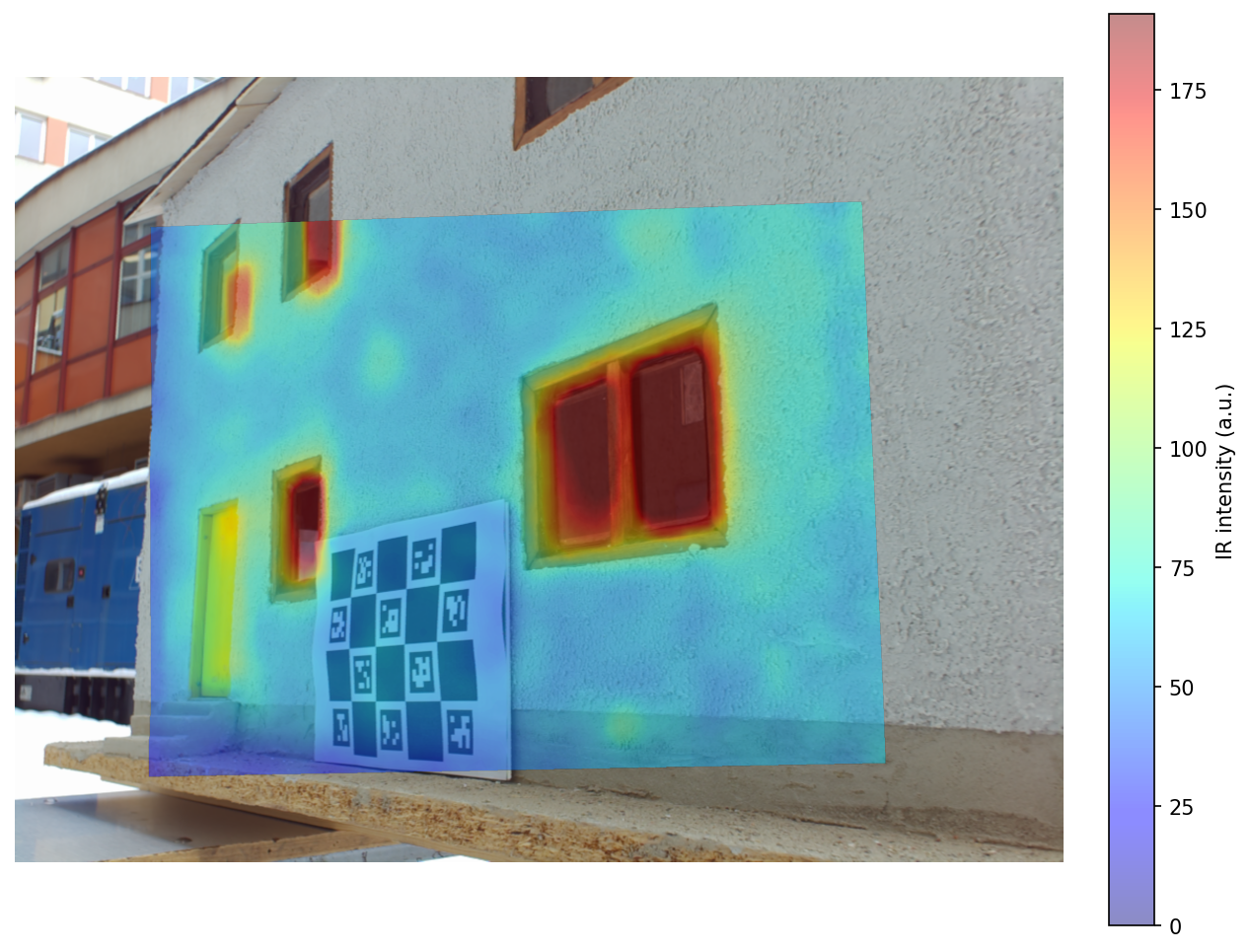}
        \small (e)
    \end{minipage}
    \hfill
    \begin{minipage}[t]{0.32\linewidth}
        \centering
        \includegraphics[width=\linewidth]{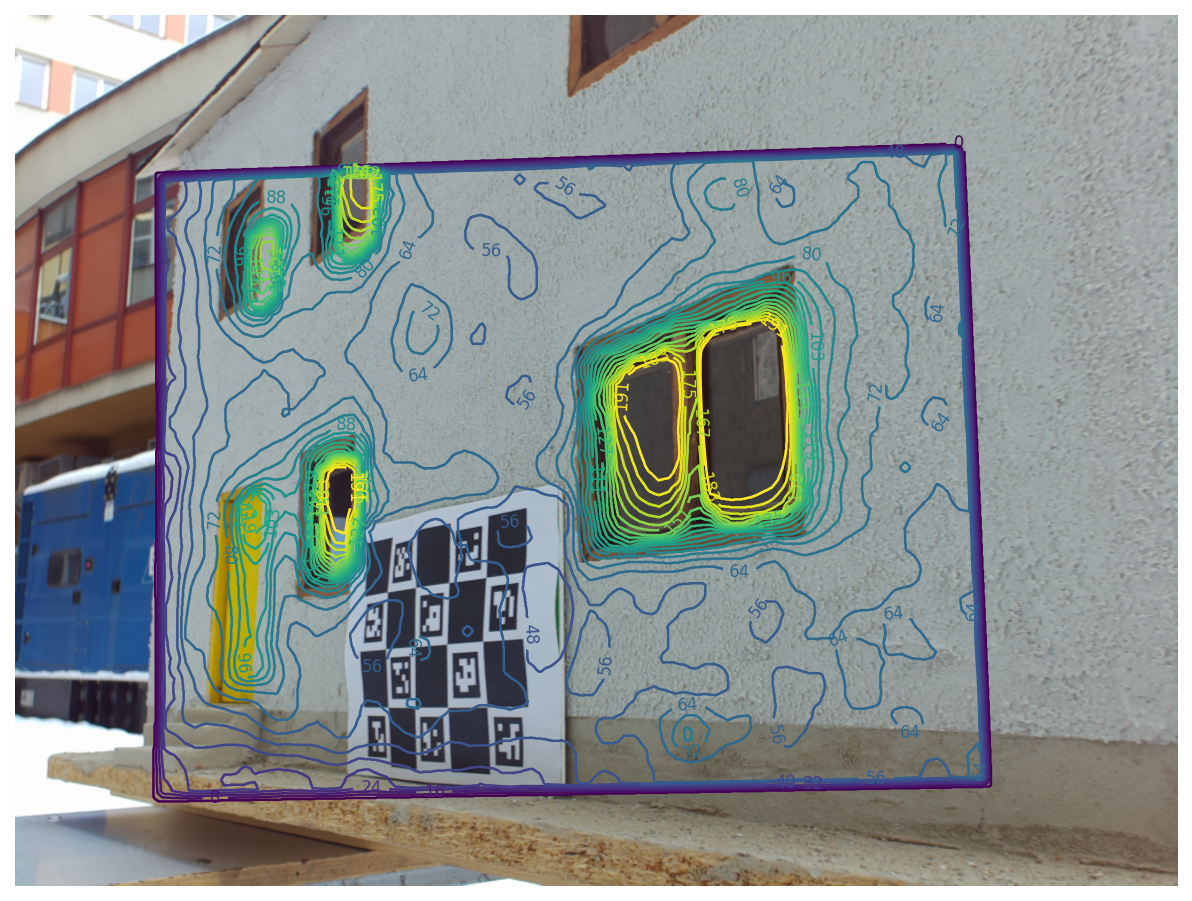}
        \small (f)
    \end{minipage}
    \caption{Comparison of RGB--TIR fusion using two approaches based on the estimated stereo calibration parameters ($R$, $T$):
    (a--c) variant using per-pixel plane-intersection depth derived from the ChArUco board pose: (a) detection and pose estimation, (b) TIR overlay on RGB, (c) TIR isolines;
    (d--f) variant using per-pixel depth estimation with DepthPro: (d) depth map illustrating the character of an outdoor scene with significant depth variation between objects, (e) TIR overlay on RGB, (f) TIR isolines.}
    \label{fig:fusion_compare}
\end{figure}

The second scenario used to verify the quality of the RGB--TIR stereo calibration involved image fusion based on per-pixel depth estimation derived from the RGB image using the DepthPro model \cite{Bochkovskii2025}. After optical radial distortion correction of both images, depth estimation was performed directly on the RGB image expressed in the geometry defined by the updated projection matrix $K_{\mathrm{RGB,new}}$. This ensured consistency between the estimated depth values and subsequent geometric transformations. An example depth map obtained from the model is shown in the form of an intensity visualisation (Fig.~\ref{fig:fusion_compare}(d)). It is worth noting the character of this depth map: for outdoor scenes with significant distance variation between objects, the depth map successfully captures both the foreground elements and the receding background structures.

Based on the estimated depth map, a corresponding three-dimensional point in the RGB camera coordinate system was computed for each RGB image pixel. These points were subsequently transformed into the coordinate system of the TIR camera using the stereo calibration parameters via Eq.~\eqref{eq:stereo_rigid}. After projection onto the TIR image plane, transformation maps were obtained and used for per-pixel remapping of the TIR image. Unlike the ChArUco-board-based approach, which assumes that all scene points lie on the calibration plane, this method imposes no planarity constraint and thus enables geometrically correct mapping of objects with arbitrary three-dimensional structure. The resulting projection of the TIR image onto the RGB image, visualised as a pseudocolour map, is presented in Fig.~\ref{fig:fusion_compare}(e).

The obtained results indicate that the use of per-pixel depth estimation further reduces parallax-related errors compared to the planar-depth approximation, particularly for scene regions that deviate from the window plane. At the same time, the quality of the TIR projection remains sensitive to the accuracy of the estimated depth map, particularly in regions with weak texture or near object boundaries. Importantly, there is a very high degree of consistency between the temperature isolines on the maps derived from the ChArUco board technique and those generated using the DepthPro model. Because the depth obtained from the ChArUco board serves as a physical reference, our experiments lead to the conclusion that in situations where physical depth measurement is not possible, applying the DepthPro model enables reliable, approximate thermal analysis using extremely low-resolution TIR cameras. Furthermore, visualisation of the TIR signal in the form of isolines overlaid on the RGB image (Fig.~\ref{fig:fusion_compare}(f)) enables a clear interpretation of thermal distributions while accounting for the true spatial structure of the scene. It is worth emphasising the very good consistency of the TIR overlay when applying both depth estimation approaches: neural (DepthPro) and homographic (ChArUco board), which is particularly evident in the temperature contour lines. A quantitative comparison of the two projection methods based on region-of-interest analysis is presented in Section~\ref{sec:roi_analysis}.

\subsection{Discussion: Quantitative Analysis in ROI Areas}
\label{sec:roi_analysis}

To complement the visual assessment of RGB--TIR data fusion quality, a quantitative analysis of thermal signal statistics was conducted within regions of interest (ROIs). The ROIs were delineated in the RGB image and correspond to characteristic scene elements, such as windows and doors, and were subsequently transferred to the TIR images using the projections obtained with both fusion methods. The ChArUco-based method employs per-pixel plane-intersection depth (Eq.~\eqref{eq:dynamic_z}), while the DepthPro-based method uses a neural per-pixel depth map. For each ROI, basic statistics of the TIR signal were computed, including the mean, median, standard deviation, and minimum and maximum values (Table~\ref{tab:roi_stats_compare}).

The comparative results reveal differences between the two projection methods across the five evaluated ROIs (Table~\ref{tab:roi_stats_compare}). For some ROIs, the DepthPro-based projection yields higher mean TIR values (e.g., $\Delta\mathrm{mean} = {+4.46}$ and ${+11.27}$~a.u.), indicating that per-pixel depth correction shifts the thermal mapping towards surface regions with greater emitted radiance. For other regions, the DepthPro variant produces lower mean TIR values (e.g., $\Delta\mathrm{mean} = {-16.41}$, ${-22.76}$, and ${-6.54}$~a.u.). This pattern is consistent with parallax-induced spatial shifts whose direction and magnitude depend on the position of each ROI relative to the stereo baseline and the local depth of the scene.

A reduction in standard deviation is observed for certain ROIs (e.g., from 26.54 to 24.43~a.u. and from 48.75 to 39.36~a.u.) in the DepthPro variant, suggesting more spatially coherent thermal projections in those regions. Conversely, the standard deviation increases slightly in other areas (e.g., from 41.34 to 45.72~a.u.), which may reflect the presence of low-intensity pixels ($\mathrm{min} = 1$~a.u.) indicative of a sensor artefact affecting the ROI boundary under per-pixel depth warping.

It should be noted that the TIR values reported in Table~\ref{tab:roi_stats_compare} correspond to 8-bit infrared intensity values and do not represent calibrated surface temperatures. No radiometric ground truth measurements were available for the present experiments; therefore, the analysis characterises the geometric consistency of the two projection methods rather than their absolute thermal accuracy. Future work should incorporate contact or radiometric reference measurements to enable a quantitative assessment of temperature mapping fidelity.

\begin{table}[H]
\centering
\small
\setlength{\tabcolsep}{3.5pt}
\renewcommand{\arraystretch}{1.15}
\caption{Statistics of the TIR signal in regions of interest (ROIs) after TIR$\rightarrow$RGB projection using two depth estimation methods: per-pixel plane-intersection depth derived from the ChArUco board pose (Eq.~\eqref{eq:dynamic_z}) and per-pixel depth estimation using DepthPro. TIR values are reported in arbitrary units (a.u.) corresponding to 8-bit infrared image intensity values in the range 0--255.}
\label{tab:roi_stats_compare}
\resizebox{\linewidth}{!}{
\begin{tabular}{l|r|rrrrr|rrrrr|r}
\hline
\multirow{2}{*}{ROI} & \multirow{2}{*}{area [\%]} &
\multicolumn{5}{c|}{ChArUco plane (TIR warped)} &
\multicolumn{5}{c|}{DepthPro (TIR warped)} &
\multirow{2}{*}{$\Delta$mean} \\
 &  & mean & median & std & min & max & mean & median & std & min & max &  \\
\hline
window\_1 & 5.27 & 163.21 & 170.00 & 47.53 &  72.00 & 240.00 & 167.67 & 172.00 & 44.07 &  68.00 & 240.00 &  +4.46 \\
window\_2 & 0.73 & 141.73 & 153.00 & 41.34 &   1.00 & 187.00 & 125.32 & 117.00 & 45.72 &   1.00 & 187.00 & -16.41 \\
window\_3 & 0.47 & 119.13 & 118.00 & 26.54 &  49.00 & 165.00 &  96.37 &  87.00 & 24.43 &  46.00 & 161.00 & -22.76 \\
window\_4 & 0.82 & 131.53 & 129.00 & 48.75 &  58.00 & 207.00 & 142.80 & 139.00 & 39.36 &  71.00 & 207.00 & +11.27 \\
door      & 0.96 &  93.98 &  95.00 & 14.24 &  58.00 & 128.00 &  87.44 &  85.00 & 16.32 &  55.00 & 128.00 &  -6.54 \\
\hline
\end{tabular}
}
\end{table}

\section{Summary and Conclusions}
\label{sec:conclusions}

This paper presented a practical and reproducible calibration framework for RGB--thermal (TIR) stereo camera systems exhibiting extreme resolution mismatch ($2028 \times 1520$ vs $80 \times 62$~px). The proposed approach was validated using a low-cost, self-assembled acquisition setup under laboratory conditions. The principal findings, corresponding to the contributions stated in Section~\ref{sec:contributions}, are summarised below.

Regarding \textbf{C1}, the use of an active OLED screen was shown to provide sufficient and repeatable thermal contrast for corner detection at $80 \times 62$~px, where passive and semi-active calibration objects failed to produce reliable detections. Dynamic switching of modality-specific patterns on the same physical surface enabled consistent geometric observations without altering the camera rig position.

Regarding \textbf{C2}, the proposed thermal corner detection algorithm---combining perspective rectification, Hessian saddle-point analysis, and Mean Shift localisation---achieved reliable checkerboard detection at very low resolution without per-frame parameter tuning or learning-based components. The fully deterministic pipeline preserved geometric interpretability while substantially improving detection robustness compared to standard OpenCV routines, which failed entirely at this resolution.

Regarding \textbf{C3}, the baseline-constrained bundle adjustment with the $T_z = 0$ constraint produced a stereo baseline of $\|T\| = 32.7$~mm, consistent with the nominal inter-camera distance of approximately 30~mm. The constrained solution achieved an overall reprojection RMS of 0.382~px (0.175~px RGB, 0.744~px TIR), while eliminating the physically implausible $T_z = 3.48$~cm estimate returned by unconstrained optimisation.

Regarding \textbf{C4}, the complete pipeline is validated on a thermally active building mock-up composed of materials with diverse thermal characteristics, demonstrating subpixel reprojection accuracy and consistent alignment of thermal isolines in TIR-to-RGB projections. Strong agreement is observed between projections obtained using calibration-based depth and DepthPro-based per-pixel depth estimation, indicating that learned depth models can be effectively combined with low-resolution TIR data for practical thermal analysis in RGB imagery.

\subsection*{Future Work}

Future research will focus on four directions. First, a natural extension is to validate the proposed pipeline on higher-resolution TIR hardware, which would confirm the generalisability of the calibration framework beyond the extreme resolution regime ($80 \times 62$~px) addressed in this work, and is expected to further improve corner localisation accuracy. Second, the integration of the calibrated RGB--TIR system with monocular depth estimation models and neural scene representations (e.g.\ NeRF, 3D Gaussian Splatting) will be investigated to enable dense thermally augmented 3D reconstruction of building envelopes. Third, radiometric calibration of the TIR sensor will be incorporated to enable quantitative surface temperature estimation and support heat transfer analysis. Fourth, field validation on real buildings under varying environmental conditions will be conducted to assess the scalability and practical applicability of the proposed calibration framework.

\section*{References}

\bibliography{references}

\begin{thebibliography}{999}

\bibitem[{International Energy Agency}(2023)]{IEA_Buildings_2023}
{International Energy Agency}.
\newblock Global Buildings and Construction Report 2023.
\newblock Technical report, International Energy Agency, Paris, France,  2023.
\newblock Accessed: 4 November 2025.

\bibitem[Wolk and Reinhart(2025)]{WolkReinhart2025}
Wolk, S.; Reinhart, C.
\newblock Semantic Building Energy Modeling: Analysis across Geospatial Scales.
\newblock {\em Building and Environment} {\bf 2025}, {\em 276},~112883.
\newblock {\url{https://doi.org/10.1016/j.buildenv.2025.112883}}.

\bibitem[Dlesk and Vach(2019)]{Dlesk2019}
Dlesk, A.; Vach, K.
\newblock Point Cloud Generation of a Building from Close Range Thermal Images.
\newblock {\em ISPRS Archives} {\bf 2019}, {\em XLII-5/W2},~29--33.
\newblock {\url{https://doi.org/10.5194/isprs-archives-XLII-5-W2-29-2019}}.

\bibitem[Zheng et~al.(2020)Zheng, Zhong, Yan, Zhao, and Wang]{Zheng2020}
Zheng, H.; Zhong, X.; Yan, J.; Zhao, L.; Wang, X.
\newblock A Thermal Performance Detection Method for Building Envelope Based on 3D Model Generated by UAV Thermal Imagery.
\newblock {\em Energies} {\bf 2020}, {\em 13},~6677.
\newblock {\url{https://doi.org/10.3390/en13246677}}.

\bibitem[Hoegner et~al.(2018)Hoegner, Abmayr, Tosic, Turzer, and Stilla]{Hoegner2018}
Hoegner, L.; Abmayr, T.; Tosic, D.; Turzer, S.; Stilla, U.
\newblock Fusion of 3D Point Clouds with TIR Images for Indoor Scene Reconstruction.
\newblock {\em The International Archives of the Photogrammetry, Remote Sensing and Spatial Information Sciences} {\bf 2018}, {\em XLII-1},~189--194.
\newblock {\url{https://doi.org/10.5194/isprs-archives-XLII-1-189-2018}}.

\bibitem[Hassan et~al.(2025)Hassan, Forest, Fink, and Mielle]{Hassan2025}
Hassan, M.; Forest, F.; Fink, O.; Mielle, M.
\newblock ThermoNeRF: A Multimodal Neural Radiance Field for Joint RGB--Thermal Novel View Synthesis of Building Facades.
\newblock {\em Advanced Engineering Informatics} {\bf 2025}, {\em 65},~103345.
\newblock {\url{https://doi.org/10.1016/j.aei.2025.103345}}.

\bibitem[Iwaszczuk and Stilla(2017)]{Iwaszczuk2017}
Iwaszczuk, D.; Stilla, U.
\newblock Camera pose refinement by matching uncertain 3D building models with thermal infrared image sequences for high quality texture extraction.
\newblock {\em Photogrammetry and Remote Sensing} {\bf 2017}, {\em 132},~33--47.
\newblock {\url{https://doi.org/10.1016/j.isprsjprs.2017.08.006}}.

\bibitem[Brenner et~al.(2023)Brenner, Reyes, Susnjak, and Barczak]{Brenner2023}
Brenner, M.; Reyes, N.H.; Susnjak, T.; Barczak, A.L.C.
\newblock RGB-D and Thermal Sensor Fusion: A Systematic Literature Review.
\newblock {\em IEEE Access} {\bf 2023}, {\em 11},~93347--93379.
\newblock {\url{https://doi.org/10.1109/ACCESS.2023.3301119}}.

\bibitem[ElSheikh et~al.(2023)ElSheikh, Abu-Nabah, Hamdan, and Tian]{ElSheikh2023}
ElSheikh, A.; Abu-Nabah, B.A.; Hamdan, M.O.; Tian, G.Y.
\newblock Infrared Camera Geometric Calibration: A Review and a Precise Thermal Radiation Checkerboard Target.
\newblock {\em Sensors} {\bf 2023}, {\em 23},~3479.
\newblock {\url{https://doi.org/10.3390/s23073479}}.

\bibitem[Zoetgnand\'e et~al.(2019)Zoetgnand\'e, Foug\`eres, Cormier, and Dillenseger]{Zoetgnande2019}
Zoetgnand\'e, Y.W.K.; Foug\`eres, A.J.; Cormier, G.; Dillenseger, J.L.
\newblock Robust Low Resolution Thermal Stereo Camera Calibration.
\newblock In Proceedings of the Eleventh International Conference on Machine Vision (ICMV 2018), Munich, Germany,  2019; Vol. 11041, {\em Proceedings of SPIE}, p. 110411D.
\newblock {\url{https://doi.org/10.1117/12.2523440}}.

\bibitem[Placht et~al.(2014)Placht, F{\"u}rsattel, Assoumou~Mengue, Hofmann, Schaller, Balda, and Angelopoulou]{Placht2014}
Placht, S.; F{\"u}rsattel, P.; Assoumou~Mengue, E.; Hofmann, H.; Schaller, C.; Balda, M.; Angelopoulou, E.
\newblock {ROCHADE}: Robust Checkerboard Advanced Detection for Camera Calibration.
\newblock In Proceedings of the Computer Vision -- ECCV 2014, Cham,  2014; Vol. 8692, {\em Lecture Notes in Computer Science}, pp. 766--779.
\newblock {\url{https://doi.org/10.1007/978-3-319-10593-2_50}}.

\bibitem[Chen et~al.(2022)Chen, Tian, He, Fu, Gu, and Wu]{Chen2022ThermalInfraredCalibration}
Chen, M.; Tian, S.; He, F.; Fu, Q.; Gu, Q.; Wu, B.
\newblock Modeling and Calibration of Active Thermal-Infrared Visual System for Industrial HMI.
\newblock {\em Electronics} {\bf 2022}, {\em 11},~1230.
\newblock {\url{https://doi.org/10.3390/electronics11081230}}.

\bibitem[Liu et~al.(2025)Liu, Chen, Yan, Cui, Xiao, Liu, and Zhang]{Liu2025}
Liu, Y.; Chen, X.; Yan, S.; Cui, Z.; Xiao, H.; Liu, Y.; Zhang, M.
\newblock ThermalGS: Dynamic 3D Thermal Reconstruction with Gaussian Splatting.
\newblock {\em Remote Sensing} {\bf 2025}, {\em 17},~335.
\newblock {\url{https://doi.org/10.3390/rs17020335}}.

\bibitem[Zhang(2000)]{Zhang2000}
Zhang, Z.
\newblock A Flexible New Technique for Camera Calibration.
\newblock {\em IEEE Transactions on Pattern Analysis and Machine Intelligence} {\bf 2000}, {\em 22},~1330--1334.
\newblock {\url{https://doi.org/10.1109/34.888718}}.

\bibitem[{OpenCV Developers}(2024)]{OpenCV_Calib3d}
{OpenCV Developers}.
\newblock OpenCV Documentation: Camera Calibration and 3D Reconstruction (calib3d Module),  2024.
\newblock Accessed: 01 December 2025.

\bibitem[Alba et~al.(2011)Alba, Barazzetti, Scaioni, Rosina, and Previtali]{Alba2011}
Alba, M.I.; Barazzetti, L.; Scaioni, M.; Rosina, E.; Previtali, M.
\newblock Mapping Infrared Data on Terrestrial Laser Scanning 3D Models of Buildings.
\newblock {\em Remote Sensing} {\bf 2011}, {\em 3},~1847--1870.
\newblock {\url{https://doi.org/10.3390/rs3091847}}.

\bibitem[Saponaro et~al.(2015)Saponaro, Sorensen, Rhein, and Kambhamettu]{Saponaro2015}
Saponaro, P.; Sorensen, S.; Rhein, S.; Kambhamettu, C.
\newblock Improving Calibration of Thermal Stereo Cameras Using Heated Calibration Board.
\newblock In Proceedings of the Proceedings of the 2015 {IEEE} International Conference on Image Processing ({ICIP}), Quebec City, QC, Canada,  2015; pp. 4718--4722.
\newblock {\url{https://doi.org/10.1109/ICIP.2015.7351702}}.

\bibitem[Roshan et~al.(2024)Roshan, Isaksson, and Pranata]{Roshan2024}
Roshan, M.C.; Isaksson, M.; Pranata, A.
\newblock A Geometric Calibration Method for Thermal Cameras Using a {ChArUco} Board.
\newblock {\em Infrared Physics \& Technology} {\bf 2024}, {\em 138},~105219.
\newblock {\url{https://doi.org/10.1016/j.infrared.2024.105219}}.

\bibitem[Vidas et~al.(2012)Vidas, Lakemond, Denman, Fookes, Sridharan, and Wark]{Vidas2012}
Vidas, S.; Lakemond, R.; Denman, S.; Fookes, C.; Sridharan, S.; Wark, T.
\newblock A Mask-Based Approach for the Geometric Calibration of Thermal-Infrared Cameras.
\newblock {\em {IEEE} Transactions on Instrumentation and Measurement} {\bf 2012}, {\em 61},~1625--1635.
\newblock {\url{https://doi.org/10.1109/TIM.2012.2182851}}.

\bibitem[Sher et~al.(2023)Sher, Xu, Chen, and Feng]{Sher2023}
Sher, B.A.; Xu, X.; Chen, G.; Feng, C.
\newblock Marker-based Extrinsic Calibration for Thermal--{RGB} Camera Pair with Different Calibration Board Materials.
\newblock In Proceedings of the Proceedings of the 40th International Symposium on Automation and Robotics in Construction (ISARC), Chennai, India,  2023; pp. 490--497.
\newblock {\url{https://doi.org/10.22260/isarc2023/0066}}.

\bibitem[Piccinelli et~al.(2024)Piccinelli, De~Rossi, Daffara, and Muradore]{Piccinelli2024}
Piccinelli, N.; De~Rossi, G.; Daffara, C.; Muradore, R.
\newblock A Passive Stereo Calibration Technique for Visible--Thermal, Low-Resolution Imaging in Remote Sensing Applications.
\newblock {\em Measurement} {\bf 2024}, {\em 231},~114647.
\newblock {\url{https://doi.org/10.1016/j.measurement.2024.114647}}.

\bibitem[Vidas et~al.(2013)Vidas, Moghadam, and Bosse]{Vidas2013}
Vidas, S.; Moghadam, P.; Bosse, M.
\newblock 3D Thermal Mapping of Building Interiors Using an RGB-D and Thermal Camera.
\newblock {\em Proceedings of the IEEE International Conference on Robotics and Automation (ICRA)} {\bf 2013}, pp. 2311--2318.
\newblock {\url{https://doi.org/10.1109/ICRA.2013.6630890}}.

\bibitem[Elias et~al.(2023)Elias, Weitkamp, and Eltner]{Elias2023}
Elias, M.; Weitkamp, A.; Eltner, A.
\newblock Multi-modal Image Matching to Colorize a SLAM Based Point Cloud with Arbitrary Data from a Thermal Camera.
\newblock {\em ISPRS Open Journal of Photogrammetry and Remote Sensing} {\bf 2023}, {\em 9},~100041.
\newblock {\url{https://doi.org/10.1016/j.ophoto.2023.100041}}.

\bibitem[Skarbek et~al.(2026)Skarbek, Salamonowicz, and Kr{\'o}l]{Skarbek2026}
Skarbek, W.; Salamonowicz, M.; Kr{\'o}l, M.
\newblock Camera Pose Revisited.
\newblock {\em Applied Sciences} {\bf 2026}, {\em 16},~2690.
\newblock {\url{https://doi.org/10.3390/app16062690}}.

\bibitem[Lu et~al.(2025)Lu, Chen, Zhu, Qin, Lu, zhang, Yan, and anke xue]{Lu2025}
Lu, R.; Chen, H.; Zhu, Z.; Qin, Y.; Lu, M.; zhang, L.; Yan, C.; anke xue.
\newblock ThermalGaussian: Thermal 3D Gaussian Splatting.
\newblock In Proceedings of the The Thirteenth International Conference on Learning Representations,  2025.

\bibitem[Beaudet(1978)]{Beaudet1978}
Beaudet, P.R.
\newblock Rotationally Invariant Image Operators.
\newblock In Proceedings of the Proceedings of the International Joint Conference on Pattern Recognition, Kyoto, Japan,  1978; pp. 579--583.

\bibitem[Garrido-Jurado et~al.(2014)Garrido-Jurado, Mu{\~n}oz-Salinas, Madrid-Cuevas, and Mar{\'i}n-Jim{\'e}nez]{GarridoJurado2014}
Garrido-Jurado, S.; Mu{\~n}oz-Salinas, R.; Madrid-Cuevas, F.J.; Mar{\'i}n-Jim{\'e}nez, M.J.
\newblock Automatic Generation and Detection of Highly Reliable Fiducial Markers Under Occlusion.
\newblock {\em Pattern Recognition} {\bf 2014}, {\em 47},~2280--2292.
\newblock {\url{https://doi.org/10.1016/j.patcog.2014.01.005}}.

\bibitem[{OpenCV Developers}(2024)]{OpenCV_Charuco}
{OpenCV Developers}.
\newblock Camera Calibration with ChArUco Boards,  2024.
\newblock Accessed: 01 December 2025.

\bibitem[Zuiderveld(1994)]{Zuiderveld1994}
Zuiderveld, K.
\newblock Contrast Limited Adaptive Histogram Equalization. In {\em Graphics Gems IV}; Heckbert, P.S., Ed.; Academic Press: San Diego, CA,  1994; pp. 474--485.

\bibitem[Otsu(1979)]{Otsu1979}
Otsu, N.
\newblock A Threshold Selection Method from Gray-Level Histograms.
\newblock {\em IEEE Transactions on Systems, Man, and Cybernetics} {\bf 1979}, {\em 9},~62--66.
\newblock {\url{https://doi.org/10.1109/TSMC.1979.4310076}}.

\bibitem[Comaniciu and Meer(2002)]{Comaniciu2002}
Comaniciu, D.; Meer, P.
\newblock Mean Shift: A Robust Approach Toward Feature Space Analysis.
\newblock {\em IEEE Transactions on Pattern Analysis and Machine Intelligence} {\bf 2002}, {\em 24},~603--619.
\newblock {\url{https://doi.org/10.1109/34.1000236}}.

\bibitem[Faugeras et~al.(2001)Faugeras, Luong, and Papadopoulo]{Faugeras1993}
Faugeras, O.; Luong, Q.T.; Papadopoulo, T.
\newblock {\em The Geometry of Multiple Images: The Laws That Govern the Formation of Multiple Images of a Scene and Some of Their Applications}; MIT Press: Cambridge, MA,  2001.

\bibitem[Hartley and Zisserman(2003)]{HartleyZisserman2003}
Hartley, R.; Zisserman, A.
\newblock {\em Multiple View Geometry in Computer Vision}, 2 ed.; Cambridge University Press: Cambridge, UK,  2003.

\bibitem[Ma et~al.(2004)Ma, Soatto, Kosecka, and Sastry]{Ma2004}
Ma, Y.; Soatto, S.; Kosecka, J.; Sastry, S.S.
\newblock {\em An Invitation to 3-D Vision: From Images to Geometric Models}; Springer: Berlin/Heidelberg, Germany,  2004.

\bibitem[Furgale et~al.(2013)Furgale, Rehder, and Siegwart]{Furgale2013}
Furgale, P.; Rehder, J.; Siegwart, R.
\newblock Unified Temporal and Spatial Calibration for Multi-Sensor Systems.
\newblock In Proceedings of the Proceedings of the IEEE/RSJ International Conference on Intelligent Robots and Systems (IROS), Tokyo, Japan,  2013; pp. 1280--1286.
\newblock {\url{https://doi.org/10.1109/IROS.2013.6696514}}.

\bibitem[Triggs et~al.(2000)Triggs, McLauchlan, Hartley, and Fitzgibbon]{Triggs2000}
Triggs, B.; McLauchlan, P.F.; Hartley, R.I.; Fitzgibbon, A.W.
\newblock Bundle Adjustment --- A Modern Synthesis. In {\em Vision Algorithms: Theory and Practice}; Triggs, B.; Zisserman, A.; Szeliski, R., Eds.; Springer: Berlin, Heidelberg,  2000; Vol. 1883, {\em Lecture Notes in Computer Science}, pp. 298--372.
\newblock {\url{https://doi.org/10.1007/3-540-44480-7_21}}.

\bibitem[Bochkovskii et~al.(2025)Bochkovskii, Delaunoy, Germain, Santos, Zhou, Richter, and Koltun]{Bochkovskii2025}
Bochkovskii, A.; Delaunoy, A.; Germain, H.; Santos, M.; Zhou, Y.; Richter, S.R.; Koltun, V.
\newblock Depth Pro: Sharp Monocular Metric Depth in Less Than a Second.
\newblock In Proceedings of the Proceedings of the 13th International Conference on Learning Representations (ICLR),  2025.
\newblock Available online: \url{https://openreview.net/forum?id=aueXfY0Clv}.

\end{thebibliography}

\typeout{get arXiv to do 4 passes: Label(s) may have changed. Rerun}
\end{document}